\documentclass[lettersize,journal]{IEEEtran}
\usepackage{amsmath,amsfonts}
\usepackage{algorithmic}
\usepackage{algorithm}
\usepackage{array}
\usepackage[caption=false,font=normalsize,labelfont=sf,textfont=sf]{subfig}
\usepackage{textcomp}
\usepackage{stfloats}
\usepackage{url}
\usepackage{verbatim}
\usepackage{graphicx}
\usepackage{cite}
\usepackage[breaklinks=true,bookmarks=true]{hyperref}

\usepackage{threeparttable}
\usepackage{tabularx}
\usepackage{color}
\usepackage{booktabs}
\usepackage{verbatim}
\usepackage{amsopn}
\usepackage{amsmath}
\usepackage{multirow}
\usepackage{colortbl}
\usepackage{makecell} 

\usepackage{colortbl,xcolor}

\usepackage{pdfpages}

\usepackage{amssymb}

\usepackage{xspace}
\makeatletter
\DeclareRobustCommand\onedot{\futurelet\@let@token\@onedot}
\def\@onedot{\ifx\@let@token.\else.\null\fi\xspace}
\def\eg{\emph{e.g}\onedot} 
\def\ie{\emph{i.e}\onedot} 
 
\def\etc{\emph{etc}\onedot} \def\vs{\emph{vs}\onedot}
 
\def\etal{\emph{et al}\onedot}
\makeatother

\newcommand{\op}[1]{\operatorname{#1}}

\usepackage{pifont}

\newcommand{\M}{HTRM-Net}

\begin{document}

\title{Repetitive Action Counting with Hybrid Temporal Relation Modeling}

\author{Kun Li, Xinge Peng, Dan Guo$^\ast$,~\IEEEmembership{Senior Member,~IEEE}, Xun Yang, Meng Wang$^\ast$~\IEEEmembership{Fellow,~IEEE}
\thanks{This work was supported by the National Key Research and Development Program of China (2022YFB4500600), the National Natural Science Foundation of China (62272144, 72188101, 62020106007 and U20A20183), the Major Project of Anhui Province (2408085J040, 202203a05020011), and the Fundamental Research Funds for the Central Universities (JZ2024HGTG0309, JZ2024AHST0337, and JZ2023YQTD0072)
(\textit{$^*$Corresponding authors: Dan Guo; Meng Wang})
}
\thanks{
K. Li and X. Peng are with the School of Computer Science and Information Engineering, Hefei University of Technology (HFUT), Hefei, 230601, China. (e-mails: kunli.hfut@gmail.com; xg.pengv@gmail.com). 
}
\thanks{
D. Guo is with the Key Laboratory of Knowledge Engineering with Big Data (HFUT), Ministry of Education, and the School of Computer Science and Information Engineering, Hefei University of Technology (HFUT), Hefei, 230601, China, and also with Institute of Artificial Intelligence, Hefei Comprehensive National Science Center, Hefei, 230026, China, and also with Anhui Zhonghuitong Technology Co., Ltd., Hefei, 230000, China. (e-mail: guodan@hfut.edu.cn).
}
\thanks{X. Yang is with the School of Information Science and Technology, University of Science and Technology of China, Hefei, 230026, China. (e-mail: xyang21@ustc.edu.cn).
}
\thanks{M. Wang is with the Key Laboratory of Knowledge Engineering with Big Data (HFUT), Ministry of Education, and the School of Computer Science and Information Engineering, Hefei University of Technology (HFUT), Hefei, 230601, China, and also with Institute of Artificial Intelligence, Hefei Comprehensive National Science Center, Hefei, 230026, China. (e-mail: eric.mengwang@gmail.com)}
}
\markboth{IEEE Transactions on Multimedia}%
{Shell \MakeLowercase{\textit{et al.}}: A Sample Article Using IEEEtran.cls for IEEE Journals}


\maketitle

\begin{abstract}
Repetitive Action Counting (RAC) aims to count the number of repetitive actions occurring in videos. 
In the real world, repetitive actions have great diversity and bring numerous challenges (\eg, viewpoint changes, non-uniform periods, and action interruptions). 
Existing methods based on the temporal self-similarity matrix (TSSM) for RAC are trapped in the bottleneck of insufficient capturing action periods when applied to complicated daily videos. 
To tackle this issue, we propose a novel method named Hybrid Temporal Relation Modeling Network (HTRM-Net) to build diverse TSSM for RAC. 
The HTRM-Net mainly consists of three key components: bi-modal temporal self-similarity matrix modeling, random matrix dropping, and local temporal context modeling. 
Specifically, we construct temporal self-similarity matrices by bi-modal (self-attention and dual-softmax) operations, yielding diverse matrix representations from the combination of row-wise and column-wise correlations. 
To further enhance matrix representations, we propose incorporating a random matrix dropping module to guide channel-wise learning of the matrix explicitly. 
After that, we inject the local temporal context of video frames and the learned matrix into temporal correlation modeling, which can make the model robust enough to cope with error-prone situations, such as action interruption. 
Finally, a multi-scale matrix fusion module is designed to aggregate temporal correlations adaptively in multi-scale matrices. 
Extensive experiments across intra- and cross-datasets demonstrate that the proposed method not only outperforms current state-of-the-art methods and but also exhibits robust capabilities in accurately counting repetitive actions in unseen action categories. Notably, our method surpasses the classical TransRAC method by 20.04\% in MAE and 22.76\% in OBO.
\end{abstract}

\begin{IEEEkeywords}
video understanding, repetitive action counting, temporal self-similarity matrix.
\end{IEEEkeywords}

\section{Introduction}\label{sec:intro}
\IEEEPARstart{R}{epetitive} Action Counting (RAC) is a challenging task\cite{dwibedi2020counting,zhang2020context,hu2022transrac,li2024repetitive} in video understanding~\cite{xia2023exploring,wang2024low,tang2021graph,li2023data,he2021dense,xu2024regennet}. 
RAC aims to count the number of repetitions of an identical action performed in the video. 
Human and animal activities in nature often exhibit repetitive motion patterns. Videos can record action variations without limitations, such as an athlete performing the same exercise repeatedly, the daily rotation of the Earth, or birds rhythmically flapping their wings. 
Counting repetitive actions in videos helps people observe the regular pattern of the action. 
Moreover, RAC can serve as an auxiliary cue for various human-centric video analysis tasks~\cite{ran2007pedestrian,guo2024benchmarking}, such as pedestrian detection~\cite{ran2007pedestrian,liu2015novel}, 3D reconstruction~\cite{li2018structure}, and heartbeat rate estimation~\cite{dwibedi2020counting,qian2024dual}. 

\begin{figure}[t]
\centering
\includegraphics[width=1.0\linewidth]{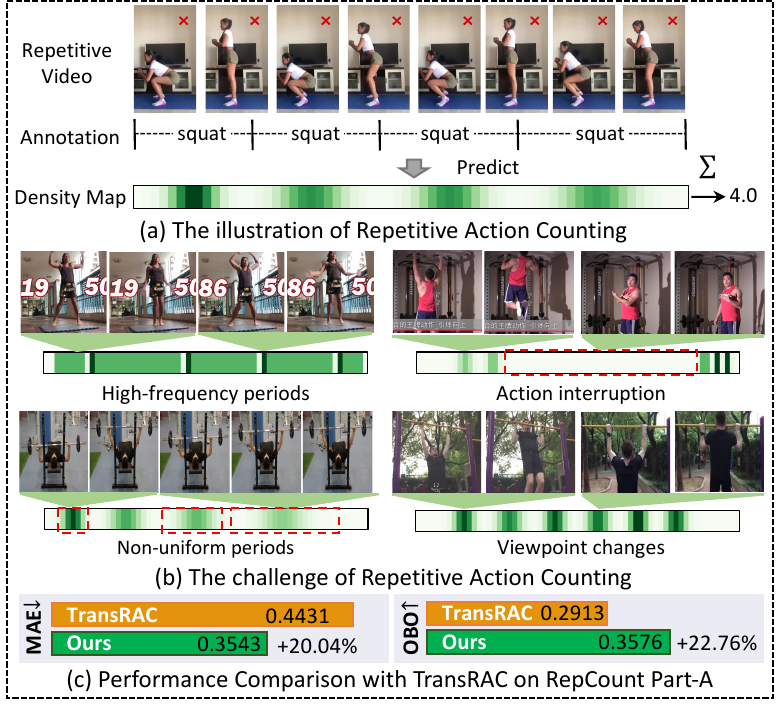}
\caption{(a) Illustration of Repetitive Action Counting (RAC), which aims to count the number of actions in a video. 
(b) The challenges in RAC include high-frequency periods, action interruption, no-uniform periods, and viewpoint changes\protect \cite{hu2022transrac}. (c) Performance comparison with classical method TransRAC~\cite{hu2022transrac}.}
\label{fig:intro}
\end{figure}

As illustrated in Figure~\ref{fig:intro} (a), given a repetitive video, a RAC model needs to predict the corresponding density map of action along the timeline for the final repetitive count prediction. 
There are four key challenges shown in Figure~\ref{fig:intro} (b). 
\textit{(1) High-frequency periods.} 
The video consists of many high-frequency actions, where the actions change so quickly and easily confuse the model. 
\textit{(2) Action interruption.} 
Physical exercise videos frequently feature action interruptions, like injuries or breaks, disrupting the original periodicity of the action cycle. 
\textit{(3) Non-uniform periods.} 
The non-uniform periods can be commonly found in strength-training actions, \eg, bench pressing, with longer exercise periods as time increases.  
Non-uniform periods emerge because of the action cycle length variation, and are always accompanied by the action frequency variation. 
\textit{(4) Viewpoint changes.} 
Videos often depict repetitive actions from various angles, demanding the model with robust temporal modeling capabilities.

Traditional methods~\cite{cutler2000robust,burghouts2006quasi,briassouli2007extraction,pogalin2008visual} targeted to address the motion periodicity for RAC, which leveraged spectral or frequency domain information of the videos. 
They adopted the motion field for periodicity analysis while ignoring context-dependency in the semantic domain. 
This makes the models easily disturbed by extreme circumstances. 
Based on hypothesis testing using a periodogram maximizer, Ran~\etal~\cite{ran2007pedestrian} examined individual pixels to determine their periodic patterns; however, this method suffered from noise sensitivity. 
Burghouts~\etal~\cite{burghouts2006quasi} proposed an online filter to identify the periodicity's position and intensity at a certain pre-tunned frequency. 
Due to the lack of large-scale datasets, these works with handcrafted features were evaluated on insufficient and over-restricted video inputs. 

\begin{figure}[t]
\centering
\includegraphics[width=1.0\linewidth]{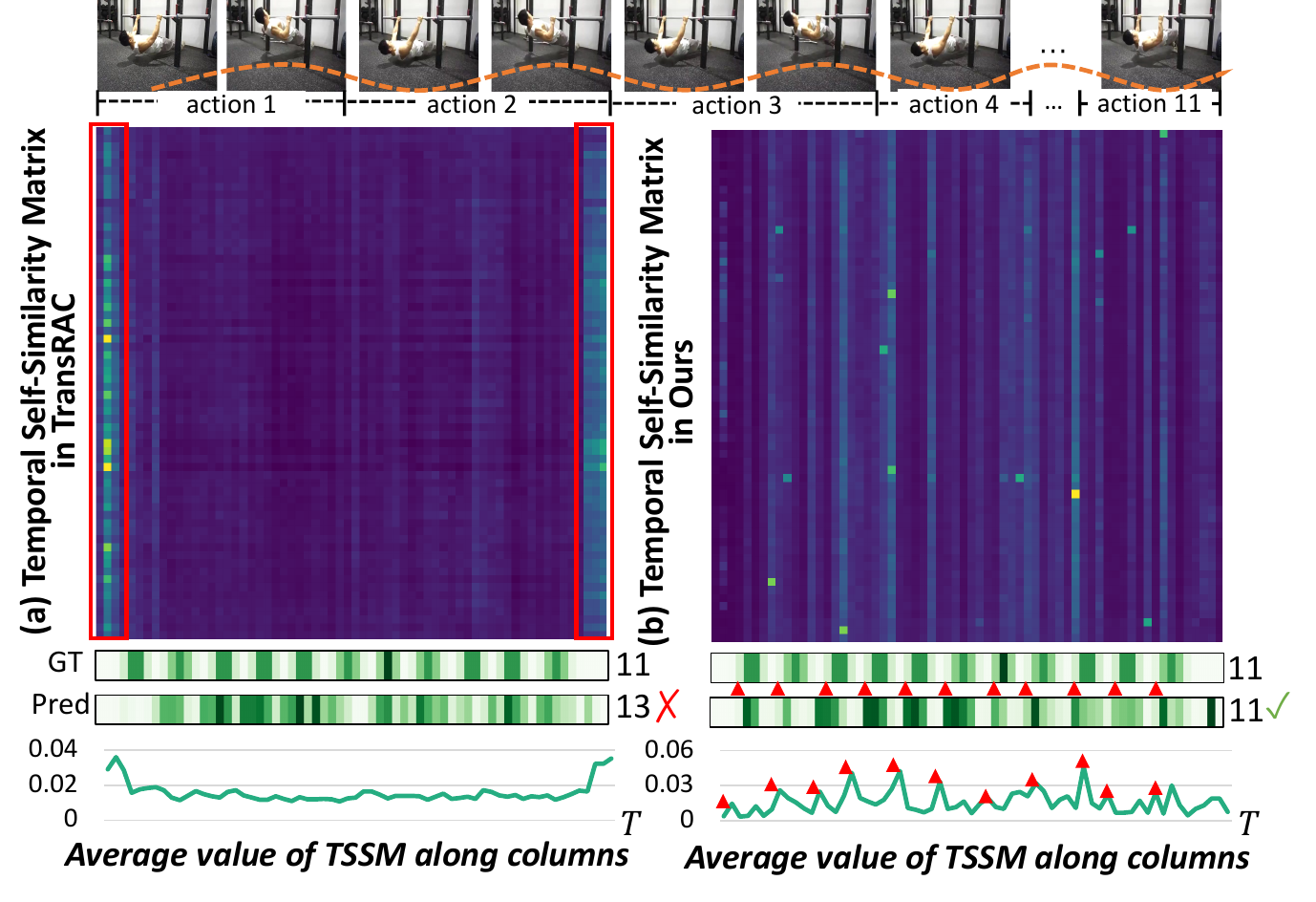}
\caption{
Illustration of (a) Temporal Self-Similarity Matrix (TSSM) in TransRAC \protect \cite{hu2022transrac} and (b) our method. 
RMD denotes the proposed Random Matrix Dropping in Sec.~\ref{sec:embedding}. 
The TSSM in our method exhibits rich temporal self-similarities corresponding to the action periods. \textcolor{red}{$\blacktriangle$} denotes the starting of each repetitive action.}
\label{fig:tsm_cmp}
\end{figure}
Nowadays, RAC has been overwhelmingly dominated by deep learning methods~\cite{levy2015live,zhang2020context,dwibedi2020counting,zhang2021repetitive,li2024repetitive}. 
The early deep learning approach~\cite{levy2015live,li2018repetitive} modeled and estimated the length of the action cycle to get the counting number. 
However, they heavily relied on the periodicity assumption and were limited to short videos, making the models unsuitable for real-world scenarios. 
Recent work studied the problem under more realistic scenarios. 
Zhang~\etal~\cite{zhang2020context} tackled the challenges in repetition counting caused by the unknown and diverse cycle lengths. 
Zhang~\etal~\cite{zhang2021repetitive} addressed video repetition estimation from a new perspective based on both vision and audio information. 
In this work, we study the pure visual RAC. 
For the visual RAC, Dwibedi~\etal~\cite{dwibedi2020counting} first proposed to utilize the \textbf{temporal self-similarity matrix (TSSM)} calculated by the negative squared Euclidean distance between visual embedding as an intermediate representation to predict action period. 
However, the TSSM calculated by the squared Euclidean distance has only one channel, which is insufficient in complex and diverse real-life scenarios. 
To address this issue, Hu~\etal~\cite{hu2022transrac} proposed TransRAC, which utilized the self-attention mechanism~\cite{vaswani2017attention} to generate representative TSSM from video embeddings. 
In addition, they introduced a more comprehensive realistic scenario dataset with different action categories and various duration lengths, named RepCount. 
The dataset provides fine-grained annotations that mark the start and end timestamps of each action cycle. 
Li~\etal~\cite{li2024repetitive} leveraged motion from the flow to guide the model focus on foreground action modeling. 
Based on the above analysis, we investigate the challenges of current research and our motivation as follows: 
\begin{itemize}

\item\textbf{Exploring diverse TSSM.} 
The TSSM matrix $ \in \mathbb{R}^{T\times T}$ has the advantage of learning the temporal similarity that lurks in multiple instances of the repetitive process and is invariant to the specific action type, which encourages model generalization in unseen action videos~\cite{dwibedi2020counting}. 
There are merely two works involving TSSM for RAC. Dwibedi~\etal~\cite{dwibedi2020counting} used Euclidean distance to compute TSSM with only one channel, which contains less correlation information. Hu~\etal~\cite{hu2022transrac} used the multi-head self-attention mechanism to address this drawback. 
As shown in Figure~\ref{fig:tsm_cmp}, we give the averaged TSSM in TransRAC~\cite{hu2022transrac}, and our method. 
It is observed that the TSSM of TransRAC tends to cluster at the beginning and end of the video (indicated in red rectangles), which lacks the precise temporal relations for RAC. 
To address this issue, we design the multi-head dual-softmax operation in both \textit{row-wise} and \textit{column-wise} measurement to build new TSSM to enhance the semantics of fine-grained temporal relations in the \textbf{\textit{spatial-wise}} of $\mathbb{R}^{T\times T}$. 
The dual-softmax operation can get the similarity of soft mutual (\textit{row-wise} and \textit{column-wise} directions) nearest neighbor matching. 
Intuitively, the multi-head dual-softmax operation generates TSSM from a new perspective that computes the bidirectional similarity, which could filter the single-side match error and get the dual match with high bidirectional similarity. 
The combination of self-attention and dual-softmax can build rich fine-grained semantics of 2D correlations. 

\item \textbf{Random TSSM Dropping.}
Although the generated bi-modal TSSM $\in \mathbb{R}^{T\times T\times C}$ contains fine-grained and precise spatial correlations, the matrix representation is still unpromising. 
A fact is that repetitive movements in RAC, such as pull-ups motion patterns, can easily lead TSSM to exhibit similar distributions in different channels and lead to correlation redundancy. 
Although the generated bi-modal TSSM contains fine-grained and precise spatial correlations, it may lack diversity in the channel dimension. 
To further enhance the self-similarity matrix representation, we consider \textbf{\textit{channel-wise}} dropping operation on TSSM. 
We randomly drop the 2D temporal matrices in the channel dimension as explicit guidance to refine the relations in matrices at the channel level. 
As a result, the generated TSSM exhibits rich temporal self-similarities corresponding to the start (\textcolor{red}{$\blacktriangle$}) of action, as in Figure~\ref{fig:tsm_cmp} (b).
The dropping operation can also improve the generalization of the model and reduce over-fitting to noise. 

\item \textbf{Enhancing local temporal context of action.} 
Although the TSSM represents a rigorous similarity calculation of the video frames in a global temporal manner, local temporal context modeling of each action is also essential for predicting the action cycle. For instance, in challenging cases, like pull-ups that consist of two sub-actions (\textit{up} and \textit{down}) with a similar motion pattern but opposite directions, the fine-grained local temporal context modeling can be built and used to eliminate the double counting error of the same sub-action. In other words, the model equipped with the local temporal context can become more robust to error-prone situations (\eg, the action interruptions in Figure~\ref{fig:intro}). 
\end{itemize}

Based on the above discussions, we propose a novel framework named \M{}. 
As shown in Figure~\ref{fig:method}, we first use the Video Swin Transformer~\cite{liu2022video} to extract multi-scale video features. 
Then, we perform hybrid temporal relation modeling. 
In bi-modal temporal self-similarity modeling, we use the self-attention and dual-softmax operations to build precise temporal self-similarity matrices, which can capture fine-grained temporal action relations. 
Subsequently, we propose a random matrix dropping module as explicit guidance to adaptively drop the self-similarity matrix in order to get diverse matrix representation (\ie, eliminating redundancy in matrix representation). In local temporal context modeling, the learned context features are used to differentiate repetitive actions. After injecting local temporal context with the self-similarity matrix, we use 3D CNNs with different kernel sizes to fuse multi-scale matrices and feed the result to the decoder for density map regression. 
As shown in Figure~\ref{fig:intro} (c), our method achieves a significant improvement over the classic TransRAC method.

In summary, the main contributions of this paper are summarized as follows: 
\begin{itemize} 
\item We propose a novel framework named \M{} for repetitive action counting, which incorporates diverse temporal self-similarity matrix and the informative local temporal context for density map generation. 

\item We discuss the importance of temporal self-similarity matrix generation and propose a bi-modal TSSM generation module and a random matrix dropping module to learn diverse temporal self-similarity matrices with fine-grained temporal relations, as well as the local temporal context to retain motion information of actions.

\item Extensive experiments validate that our method outperforms the state-of-the-art method by 9.82\% in terms of MAE on the RepCount-A dataset. Cross-dataset experiments validate the robustness of our method on unseen action categories. 
\end{itemize}

\section{Related Work}
In this section, we first review related works on the task of repetitive action counting. We focus on representative learning of temporal self-similarity matrix (TSSM) modeling for repetitive action counting. Therefore, we introduce works related to TSSM applied to video understanding tasks. 

\subsection{Repetitive Action Counting} 
Different from crowd counting~\cite{gong2024adaptive,zhu2024find,liang2023focal,guo2019dadnet}, which counts humans in a crowded image or video, repetitive action counting~\cite{levy2015live,runia2018real,dwibedi2020counting,hu2022transrac,li2024repetitive} aims to count the number of repetitive actions in the video. 
Early works typically focused on action periodicity analysis~\cite{chetverikov2006motion} or motion variation analysis~\cite{belongie2006structure}. 
Initially, these studies learned the motion field and converted it into one-dimensional signals.
Subsequently, various traditional methods were employed for analysis, including Fourier analysis~\cite{tsai1994cyclic,briassouli2007extraction}, peak detection~\cite{thangali2005periodic}, singular value decomposition~\cite{chetverikov2006motion}, and classification~\cite{davis2000categorical}. In summary, these traditional methods are limited by assuming action frequency as a distinguishable peak in time-frequency patterns.

Recently, various data-driven approaches and datasets have been proposed for RAC. 
Levy~\etal~\cite{levy2015live} collected the YTsegments dataset, which contains 100 videos with the activities of human exercising, cooking, building, \etc. 
However, it only contains stationary repetitions, which limits the generalizability of the model.  
Levy~\etal~\cite{levy2015live} formulated RAC as a special count number classification problem, and combined a region of interest detection for robust action counting. 
In addition, they also collected the YTsegments dataset, which contains 100 videos with the activities of human exercising, cooking, building \etc.  
However, the YTsegments dataset only contains stationary repetitions, which limited the generalizability of the model.  
To address this limitation, 
\cite{runia2018real} collected the QUVA Repetition dataset containing non-stationary repetitions. 
However, YTsegments and QUVA Repetition datasets contain only 100 videos, this potentially hampers the model's counting accuracy in real scenarios. 
\cite{dwibedi2020counting} introduced a large-scale dataset named Countix from the subset of the Kinetics 400 dataset~\cite{kay2017kinetics}. 
This dataset consists of over 8,000 videos, which is based on the assumption that the motion is continuous and uniformly distributed. 
Different from these methods, Zhang~\etal~\cite{zhang2021repetitive} proposed to incorporate sound into the repetition counting, while repetitive actions often contain noisy background noise. 
To fill the gap between synthetic data and real data, Zhang~\etal~
\cite{zhang2020context} collected a dataset named UCFRep. 
However, the above-mentioned datasets are annotated with a total number of repetitive actions. 

In real-life scenarios, the repetitive action is usually interrupted by other incidents. 
Therefore, the coarse-grained annotations that merely provide the total counts of these repetitive actions are insufficient for supporting fine-grained RAC. 
To fascinate the fine-grained RAC, Hu~\etal~\cite{hu2022transrac} proposed the RepCount dataset with the annotation of each cycle for the first time and introduced density map regression to counting actions. The density map preserved the temporal distribution of action in video and is suitable for arbitrary video lengths. 
Li~\etal~\cite{li2024repetitive} leveraged motion from the flow to guide the model focus on foreground action modeling for better density map generation.  
In this work, we aim to address the repetitive action counting involved in daily activities that are non-static and non-stationary based on density map prediction.

\begin{figure*}[t]
\centering
\includegraphics[width=1.0\linewidth]{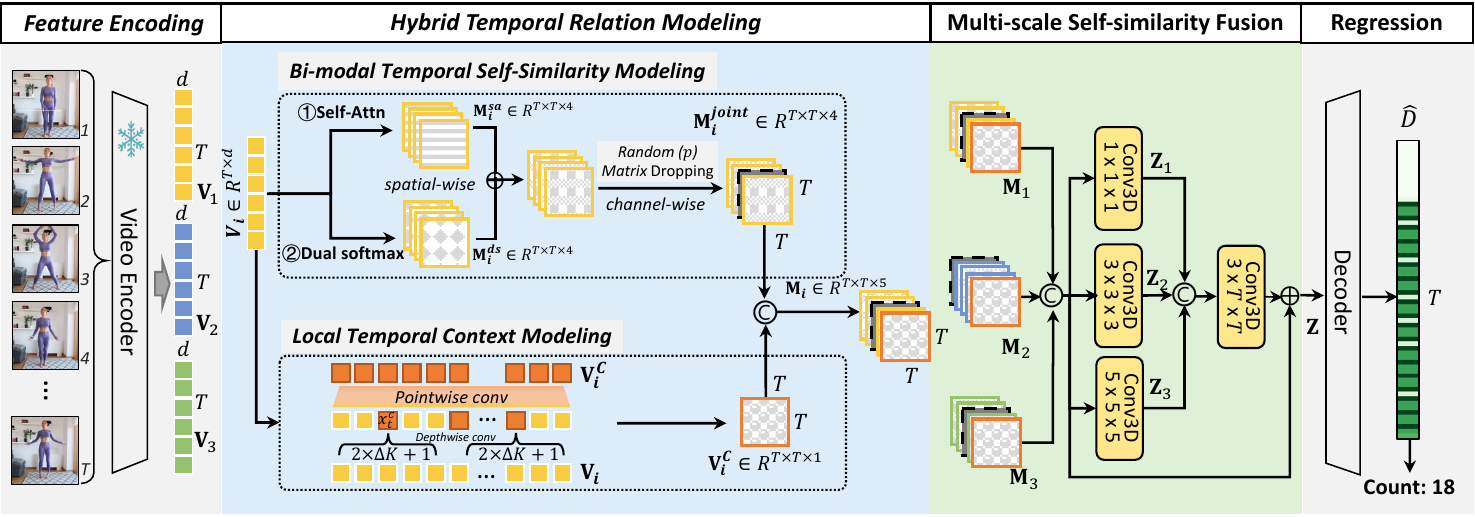}
\caption{Overview of the proposed Hybrid Temporal Relation Modeling Network (\M). 
First, we extract multi-scale video features $\mathbf{V}_i, i\in\{1,2,3\}$, which are used to generate the temporal self-similarity matrix in a bi-modal manner and build the local temporal context concurrently. 
Then, in the bi-modal temporal self-similarity modeling, we use multi-head self-attention and dual-softmax operations to build fine-grained temporal relations in a spatial-wise manner. 
Subsequently, the random matrix dropping (RMD) is applied to the bi-modal matrices to build diverse matrix representation in a channel-wise manner through the random matrix dropping strategy.  
Meanwhile, we inject the local temporal context into the temporal self-similarity matrix to prevent error-prone situations. 
Additionally, we design a multi-scale self-similarity fusion module to fuse the temporal self-similarity matrix in each scale.
Finally, we fuse the multi-scale matrices and use the decoder to predict the target density map $\hat D$. 
$T$ represents number of frames while $d$ denotes number of channels.
}
\label{fig:method}
\end{figure*}

\subsection{Temporal Self-Similarity Matrix in Video Understanding}
The temporal self-similarity matrix (TSSM) aims to measure temporal similarity or correlation between two video sequences and has been widely used in various video understanding tasks~\cite{nam2021zero,wang2024eulermormer,qian2024cluster,wang2024frequency}, such as action recognition, video copy localization~\cite{he2023transvcl}, video grounding~\cite{liu2024conditional,nam2021zero,li2021proposal,kim2023language}, and generic action boundary detection~\cite{du2022fast}. 
In addition, temporal similarity explored context dependency in the temporal domain due to its robustness in calculating similarity with different scenarios. 
It inherently has the properties of internal consistency within actions and external discrepancy across actions. 
The temporal self-similarity matrix is employed to view independent action recognition for human action recognition~\cite{junejo2010view,sun2015exploring,kwon2021learning}.
Specifically, Junejo~\etal~\cite{junejo2010view} computed the temporal similarity from different image features that possess similar properties.
Kwon~\etal~\cite{kwon2021learning} utilized spatio-temporal similarity to represent each local region as similarities to its neighbors in space and time for video action recognition. 
In video copy localization, He~\etal~\cite{he2023transvcl} utilized the temporal similarity matrix generated from a differentiable softmax matching layer to present more discriminative copied patterns. 
For generic action boundary detection, Du~\etal~\cite{du2022fast} calculated the cosine similarity between adjacent frames while transferring the boundary detection task into the change point detection based on the similarity. 
As for the weakly supervised video grounding, the multi-instance learning-based method \cite{zhang2020counterfactual} used the temporal similarity scores to maximize scores between positive samples and minimize scores between negative samples.
Tan~\etal\cite{tan2021logan} utilized similarity scores between pairs for segment features to build the pseudo-temporal proposals for later video grounding. 
For zero-shot video grounding, \cite{nam2021zero,kim2023language} employed the temporal self-similarity matrix to find temporal event proposals by calculating the cosine similarity score between pairs of segment features. 

For RAC, Dwibedi~\etal~\cite{dwibedi2020counting} constructed TSSM by computing squared Euclidean distance between all pairwise features. 
However, this hand-crafted manner only builds one channel feature. 
To improve the generalization of TSSM, Hu~\etal~\cite{hu2022transrac} employed self-attention~\cite{vaswani2017attention} to generate a more robust temporal similarity matrix. In this work, we focus on modeling the temporal self-similarity matrix and propose a hybrid temporal relation modeling network that incorporates diverse temporal self-similarity matrix and informative local temporal context for repetitive action counting.

\section{Methodology}
\subsection{Overview}
RAC aims to count the repetitive actions in videos. In this study, we formulate this task as a temporal density map regression problem. 
Given a repetitive video $\mathbf{X}$ with $T$ frames, the goal is to learn a model $\mathcal{F}$ to predict the corresponding density map $\hat{D}\in \mathbb{R}^{T}$: 
\begin{equation}
\hat{D} = \mathcal{F}(\mathbf{X};\Theta),
\end{equation}
where $\Theta$ is the parameter collection of model $\mathcal{F}$. 
Here, we propose a novel hybrid temporal modeling network (HTRM-Net) to build diverse TSSM for RAC.
As shown in Figure~\ref{fig:method}, \M{} is composed of four stages: feature encoding, hybrid temporal relation modeling, multi-scale self-similarity fusion, and density map regression.  
We first extract multi-scale video features $\mathbf{V}_i\in \mathbb{R}^{T\times d}$. 
Then, each $\mathbf{V}_i$ is fed to dual-path temporal self-similarity learning to build rich self-similarity matrix $\mathbf{M}_i$ to capture repetitive patterns. 
Subsequently, multi-scale self-similarity fusion is used to aggregate the matrix $\mathbf{M}_i$ at each scale. 
Finally, the feature decoder is utilized to predict the density map $\hat{D}\in \mathbb{R}^{T}$, which contributes to the total count calculation. 

\subsection{Feature Encoding}
Let the input video as $\mathbf{X} \in \{x_i\}_{i=1}^{T}$, where $x_i \in  \mathbb{R}^{H_0\times W_0\times 3}$ is the $i$-th frame, $H_0$ and $W_0$ is the width and height of each frame, and $T$ is number of total frames. 
Considering that repetitive actions typically have different periods, we sample the video using sliding windows of different sizes to construct multi-scale features. 
Specifically, we used three sliding windows of sizes 1, 4, and 8 with a temporal overlap ratio of 0.5. 
In this way, the original video is extracted into three scales: single frame ($\mathbf{X}_{1} \in \mathbb{R}^{T\times 1\times H_0\times W_0\times 3}$), 4 frames ($\mathbf{X}_{2}  \in  \mathbb{R}^{T/2\times 4\times H_0\times W_0\times 3}$), and 8 frames ($\mathbf{X}_{3} \in \mathbb{R}^{T/4\times 8\times H_0\times W_0\times 3}$). 
The Video Swin Transformer is used to extract video features. 
For each scale feature $\mathbf{X}_{i}$, all the clip features are concatenated in the temporal dimension to build the output feature $\mathbf{X}_{i}^{\prime}$ is of $T \times 7 \times 7 \times 768$, where $i \in \{1, 2, 3\}$. 
For each scale feature $\mathbf{X}_{i}^{\prime}$, we use a 3D convolution $\op{Conv3D}$ with the filter size of $3 \times  3 \times 512$ followed by the batch normalization~\cite{ioffe2015batch} and ReLU activation to incorporate more temporal contexts into each clip feature. 
To reduce model parameters, we apply global 3D Max-pooling on the spatiotemporal features to fold spatial dimensions to obtain the video feature $\mathbf{V}_i\in\mathbb{R}^{T\times d}$: 
\begin{equation}
\mathbf{V}_{i} = \op{MaxPool3D}(\mathbf{X}_{i}^{\prime}) \in \mathbb{R}^{T\times d}, i\in \{1,2,3\}, 
\end{equation}
where the kernel size of $\op{MaxPool3D}$ is $1 \times 7 \times 7$. 

\subsection{Hybrid Temporal Relation Modeling}~\label{sec:embedding}
To model the temporal relation for accurate RAC, we perform hybrid temporal self-similarity learning by two distinct paths: bi-modal temporal self-similarity modeling to generate diverse TSMM $\mathbf{M}^{joint}_{i}$, and local temporal context modeling to capture the local contextual feature $\mathbf{V}_i^{C}$.

\subsubsection{Bi-modal Temporal Self-Similarity Modeling}

In repetitive action counting, the temporal self-similarity matrix aims to capture the periodic pattern of actions. As discussed in Sec.~\ref{sec:intro}, in order to alleviate the insufficient repeatability of TSSM, we attempt to construct rich temporal self-similarity matrix to capture action periods. 
Here, we introduce the generation of bi-modal temporal self-similarity matrices and the random matrix dropping module. 

\noindent\textbf{Bi-modal Temporal Self-Similarity Matrices.}
Firstly, we generate the temporal self-similarity matrix using two different approaches: the multi-head self-attention and multi-head dual-softmax operations. 
The matrix $h_{t}^{sa}$ is constructed by correlation expression $C_{i}^{sa}=< \mathbf{V}_{i}^{Q}, (\mathbf{V}_{i}^{K})^{\intercal} >$, followed by the scale and softmax operations. $\mathbf{V}_{i}^{Q}$ and $\mathbf{V}_{i}^{K}$ is obtained by two linear layers on the input feature $\mathbf{V}_{i}$ separately. 
The process can be formulated as: 
\begin{equation}
\mathbf{h}_h^{sa} = \op{softmax}(\mathbf{V}_{i,h}^{Q} (\mathbf{V}_{i,h}^{K})^{\intercal} / {\sqrt{d}})_{row},
\label{eq:attn}
\end{equation} 
\begin{equation}
\mathbf{M}_i^{sa} = \mathbf{W}_{sa}[\mathbf{h}_1^{sa};\mathbf{h}_2^{sa};\mathbf{h}_3^{sa};\ldots;\mathbf{h}_{H}^{sa}],
\label{eq:m_sa}
\end{equation} 
where $\mathbf{W}_{sa} \in \mathbb{R}^{d\times d}$ is a trainable parameter, and $[;]$ is the concatenate operation. This operation could obtain the global context and capture the long-range dependencies. Here, we adopt 4 heads (\ie, $H$=4) to build the TSSM, thereby $\mathbf{M}_i^{sa}\in\mathbb{R}^{T\times T\times 4}$. 

However, as discussed in Sec.~\ref{sec:intro}, it is found that $\mathbf{M}_{i}^{sa}$ is insufficient to represent rich self-similarity of temporal information. 
Inspired by \cite{he2023transvcl}, we design the multi-head dual-softmax operation, both in the sense of softer decision and better differentiability properties, that can be applied to similarity matrix generation with fine-grained and precise temporal correlation. 
Firstly, we construct the correlation expression $C_{i}^{ds}=<\mathbf{V}_{i}^{Q}, (\mathbf{V}_{i}^{K})^{\intercal} >$. 
Subsequently, the softmax operates on the \textit{row} and \textit{column} dimensions, and the Hadamard product is used to build the similarity matrix $\mathbf{M}_{i}^{ds}$: 
\begin{equation}
\mathbf{h}_h^{ds}=\op{softmax}(\mathbf{V}_{i,h}^{Q} (\mathbf{V}_{i,h}^{K})^{\intercal})_{col}\odot \op{softmax}(\mathbf{V}_{i,h}^{Q} (\mathbf{V}_{i,h}^{K})^{\intercal})_{row},
\label{eq:dual1}
\end{equation}
\begin{equation}
\mathbf{M}_i^{ds} = \mathbf{W}_{ds}[\mathbf{h}_1^{ds};\mathbf{h}_2^{ds};\mathbf{h}_3^{ds};\ldots,\mathbf{h}_{H}^{ds}],
\label{eq:m_ds}
\end{equation}
where $\odot$ represents the element-wise multiplication, and $\mathbf{W}_{ds} \in \mathbb{R}^{d\times d}$ is a trainable parameter. 
The multi-head dual-softmax operation can obtain the similarity of soft mutual nearest neighbor matching. It can also filter the case with a large gap between the similarity score in two cross directions. 

After generating the similarity matrices $\mathbf{M}^{sa}_i$ and $\mathbf{M}^{ds}_i$, we build the final similarity matrix $\mathbf{M}_i^{joint}\in \mathbb{R}^{T\times T\times 4}$: 
\begin{equation}
\mathbf{M}_i^{joint} = \mathbf{M}_i^{sa} \oplus \mathbf{M}_i^{ds},
\label{eq:add}
\end{equation}
where $\oplus$ denotes the element-wise addition. 
The addition of corresponding values from each matrix can eliminate the influence of extreme values in one single matrix. 
Additionally, the score disparity between high and low similarity increased, yielding more convincing results. Through the bi-modal strategy, fine-grained temporal relations are further enhanced spatially. 

\noindent\textbf{Random Matrix Dropping.} 
As stated in Sec.~\ref{sec:intro}, we discovered that the generated self-similarity matrices represent the temporally similar distribution between channels, especially the time periods of high similarity, which leads to redundant responses in the density map around the same time period. 
To yield diverse matrix representation, we propose \textit{channel-wise} matrix learning. 
Inspired by the Dropout operation~\cite{srivastava2014dropout}, can prevent over-fitting of the model, we propose the Random Matrix Dropping (RMD) operation to enrich the representability of self-similarity matrix. 
The mechanism of Dropout is to randomly set some elements to zero with a probability $p$ using samples from a Bernoulli distribution. 
In contrast, RMD adopts multi-channel similarity matrix as an input and treats each matrix as a whole. 
Specifically, for $\mathbf{M}_{i}$ in each scale, RMD randomly drops one matrix in $\mathbf{M}_{i}^{joint}$ by setting it to zero entirely. The drop operation is solely taken in the training process and is performed with a probability $p$. 

\subsubsection{Local Temporal Context Modeling}
Given that the temporal self-similarity matrix only captures the correlation between video frames, and temporal context helps to model short-term motion and enables the model to distinguish between the frames with similar motion patterns (\eg, hands moving up or down while exercising). 
As shown in Figure~\ref{fig:method}, we leverage 1D separable convolution~\cite{chollet2017xception} to model local temporal context. 
Initially, for each time $t$, we employ the \textit{depthwise} convolution with the windows size of ($2 \times \Delta K + 1$) to cover the adjacent $\Delta K$ frames before and behind current time to capture local temporal context $x_{t}^{\prime}$. Subsequently, we leverage \textit{pointwise} convolution with $1 \times  1$ kernel to reduce the feature dimension from $d$ to $T$ to get feature $x_{t}^{c}$. 
The calculation can be formulated as follows:
\begin{equation}
\begin{aligned}
x_t^{c} &= \text{SeparableConv}(x_t, \Delta K) \\
&=\left\{
\begin{array}{ll}
    x_t^{\prime} =   \text{Conv1D}(x_{t-\Delta K}, ..., x_t, ..., x_{t+\Delta K})\vert_{\text{depthwise}}; \\
    x_t^{c} =   \text{Conv1D}(x_t^{\prime})\vert_{\text{pointwise}}.
\end{array}
\right.
\end{aligned}
\label{eq:conv1d}
\end{equation}
Therefore, we can get the local temporal context feature $\mathbf{V}_{i}^{C} = \{x_1^{c}, x_2^{c}, ..., x_T^{c}\} \in \mathbb{R}^{T\times T\times 1}$. 
The hyper-parameter $k_{conv}$ will be discussed in Sec.~\ref{sec:abl} Ablation Study. 
Finally, we concatenate it to temporal 
self-similarity matrix $\mathbf{M}_i^{joint}$ to build the enhanced feature $\mathbf{M}_i \in \mathbb{R}^{T\times T\times5}$.

\subsection{Multi-scale Self-Similarity Fusion}
After obtaining the enhanced features ${\mathbf{M}_1, \mathbf{M}_2, \mathbf{M}_3}$ at all scales, we design a temporal self-similarity matrix fusion module to fuse these features. Considering that multi-scale matrices reflect different temporal self-similarity in each scale, we design a multi-scale self-similarity fusion module to aggregate these enhanced features. 
As shown in Figure~\ref{fig:method}, we first perform 3D CNNs with the kernel size of $\{1 \times 1 \times 1, 3 \times  3 \times  3, 5 \times  5 \times  5\}$ to fuse the feature and build the output $\mathbf{Z}_i\in \mathbb{R}^{T\times T\times 5}$, $i\in \{1,2,3\}$. 
After that, we concatenate $\mathbf{Z}_i, i\in \{1,2,3\}$ together and use a 3D CNN with the kernel size of $3\times T\times T$ to reduce model parameters and add the original $\mathbf{M}_{i}, i\in \{1,2,3\}$ to it. 
Finally, we obtain the diverse matrix feature $\mathbf{Z}\in \mathbb{R}^{T\times T \times 15}$.

\subsection{Density Map Regression and Optimization}\label{sec:predicotr}
\subsubsection{Density Map Regression} 
We leverage one transformer encoder layer~\cite{vaswani2017attention} to decode the diverse matrix feature $\mathbf{Z} \in \mathbb{R}^{T\times T\times 15}$, and an MLP to predict density map $\hat{D}\in\mathbb{R}^{T}$: 
\begin{equation}
\mathbf{O}=\op{TransEnco}(\op{Flatten}(\mathbf{Z})),
\end{equation}
\begin{equation}
\hat{D}=\op{MLP}(\mathbf{O}),
\label{eq:feature_O}
\end{equation}
where $\op{TransEnco}$ denotes the vanilla transformer encoder layer, and $\mathbf{O}\in\mathbb{R}^{T\times d}$. 
$\op{Flatten}$ operation converts feature $\mathbf{Z}$ to $\mathbf{Z}^{\prime} \in \mathbb{R}^{T\times (T\times 15)}$. 
$\op{MLP}$ contains two fully connected layers with a ReLU layer. 
In the predicted density map $\hat D = [\hat d_{1}, \hat d_{2}, \ldots, \hat d_{n}]$, $\hat d_{i}$ represents the density of counts at $i$-th frame. 
Perform a summation operation on $\hat D$ can obtain the total count of the video. 
The value of density $\hat D = [\hat d_{1}, \hat d_{2}, \ldots, \hat d_{n}]$ indicates global information of the whole period, which represents the distribution of the action cycle. It also contains the local information on the frame’s position in the local cycle. 

\subsubsection{Model Optimization and Inference}\label{sec:loss}
Following TransRAC, we use the MSE (Mean Squared Error) loss between the ground truth density map $D$ and the predicted density map $\hat D$. 
Specifically, the optimization function is applied to every $i$-th frame instead of the total count of the video. 
The MSE (Mean Squared Error) loss $\mathcal{L}$ can be formulated as: 
\begin{equation}
\mathcal{L} = \frac{1}{N} \sum_{i=1}^{N} || D_i - \hat{D}_i  || ^ 2_F ,
\end{equation}
where $N$ denotes the total number of the train samples, and $|| \cdot ||^2_{F}$ denotes the Frobenius norm of a matrix. 
Note that the repetitive action category information is not used in the model.  
During training and inference, we uniformly sample $T = 64$ frames from each video sample in the dataset. 
If the video frames are less than 64, we pad it to 64 frames. In the output density map $\hat D=[\hat d_{1}, \hat d_{2}, \ldots, \hat d_{n}]$, $\hat d_{i}$ represents the value of probability density at $i$-th frame. We sum up $\hat{D}$ as the predicted counts. 

\section{Experiments}
In this section, we first introduce the experimental setup on three public datasets: RepCount, UCFRep, and QUVA. Subsequently, we make comparisons with state-of-the-art methods and conduct ablation studies to validate the effectiveness of the proposed \M{}. Finally, we provide qualitative visualization and analysis of the prediction results. 

\subsection{Experimental Setup}
\subsubsection{Datasets} 
In this paper, we adopt three datasets to evaluate the performance of the proposed model:
\textbf{(1) RepCount}~\cite{hu2022transrac} consists two parts, \ie, Part-A and Part-B. 
\textbf{Part-A} consists of 1,041 videos with variable length and anomaly cases (\eg, action interruptions, and non-uniform periods) from YouTube, which is a challenging dataset.  
\textbf{Part-B} is collected in real life, which contains the fitness exercises (\eg, \textit{sit up} and \textit{pull ups}) performed by student volunteers. 
It contains 410 videos with an average duration of 28.53 seconds. 
Since the Part-B dataset has not been released publicly\footnote{\url{https://github.com/SvipRepetitionCounting/TransRAC/issues/44}}, we only evaluate our model on the Part-A dataset. 
\textbf{(2) UCFRep} is collected by Zhang~\etal~\cite{zhang2020context}, which is derived from the subset of the UCF101 dataset~\cite{soomro2012ucf101}. 
This dataset contains 526 videos in 23 different repetition action classes with an average duration of 8.15 seconds.
There are 421 and 105 videos in the training and test set, respectively. 
Following the protocol in~\cite{hu2022transrac}, we also use this dataset for cross-dataset evaluation. 
\textbf{(3) QUVA Repetition} is collected by Runia~\etal~\cite{runia2018real}, which only contains 100 non-stationary videos for evaluation. 
Limited by the size of this dataset, previous works~\cite{runia2018real,zhang2020context} typically generated synthetic data to train the model, but there is a non-negligible domain gap between synthetic data and real data. Thus, we only use it for cross-dataset evaluation. 
Considering that the UCFRep and QUVA datasets are small, we use these datasets for cross-dataset evaluation as done in TransRAC~\cite{hu2022transrac}.
Although there are some similar datasets, such as Countix~\cite{dwibedi2020counting}, Countix-AV~\cite{zhang2021repetitive} and YTsegments~\cite{levy2015live}. We find them inadequate for evaluating our model because Countix and Countix-AV datasets lack fine-grained annotations (\ie, the start and end times of each action), while YTsegments only contains 100 video clips.  
Therefore, we do not use the Countix and YTsegments datasets in our work. 

\subsubsection{Evaluation Metrics} 
Following the convention of existing methods~\cite{zhang2020context,zhang2021repetitive,hu2022transrac} for RAC, we evaluate the proposed model with two metrics, \ie, Mean Absolute Error (\textbf{MAE}) and Off-By-One (\textbf{OBO}) accuracy. 
\textbf{MAE} denotes the normalized absolute error between the ground truth count and the predicted count. 
The lower the MAE, the more accurate the predicted results are. 
\textbf{OBO} represents the correct rate of repetition count over the entire dataset. 
If the predicted count is within one count of the ground truth, we deem this video to be counted correctly. Otherwise, it is counting error. 
The higher the OBO, the more accurate the counting of action. 
Specifically, MAE and OBO are formulated as follows:
\begin{equation}
MAE=\frac{1}{N}\sum_{i=1}^{N}\frac{\left | c_{i} - \hat{c_{i}}  \right | }{\tilde{c_{i}}},
\end{equation}
\begin{equation}
OBO=\frac{1}{N}\sum_{i=1}^{N}\left [\left | c_{i} -  \hat{c_{i}} \right |\le 1 \right ], 
\end{equation}
where $c_i$ and $\hat{c_{i}}$ are the ground truth and predicted count of $i$-th video, respectively. $N$ is the number of test videos. 
$\hat{c_{i}}$ is obtained by summing over the predicted density map $\hat D_i$ across each timestamps.

\subsubsection{Implementation Details} 
We use the Video Swin Transformer tiny version~\cite{liu2022video} as the video encoder, which is pre-trained on the Kinetics-400 dataset~\cite{kay2017kinetics}. 
Note that we do not fine-tune the Video Swin Transformer model during training. 
The input image is resized to 224$\times$224. 
The hidden dimension $c$ is set to 512. 
The video is sampled with 64 frames. 
The parameters of each module can be found in the methodology. 
Additionally, we use the Adam optimizer~\cite{kingma2014adam} to optimize the network with the initial learning rate of $1\times 10^{-4}$ and batch size of 32. The proposed \M{} is implemented in PyTorch~\cite{NEURIPS2019_9015}.

\begin{table}[t!]
\centering
\tabcolsep 8pt
\caption{The data statistics of close-set setting and open-set setting of RepCount Part-A dataset. \#classes denotes the number of action categories.}
\renewcommand\arraystretch{1.0}
\resizebox{1.0\linewidth}{!}{
\begin{tabular}{c|cc|cc}
\toprule
\multirow{2}{*}{Split} & \multicolumn{2}{c|}{Regular setting} & \multicolumn{2}{c}{Open-set setting} \\ \cline{2-5} 
& \#Videos   & \#classes  & \#Videos & \# classes \\ \hline
training & 758 & 10  & 646 & 7 \\
validation   & 131 & 9   & 128 & 1  \\
test  & 152 & 9   & 250 & 2  \\ \bottomrule 
\end{tabular}}
\label{tab:dataset}
\end{table}

\begin{table}[t]
\centering
\tabcolsep 15pt
\caption{Performance comparison on RepCount-A with existing methods under regular and open-set settings using RGB features. 
$^\star$ denotes the model with only RGB features. 
The best results are highlighted in \textbf{bold}.
}
\renewcommand\arraystretch{1.0}
\resizebox{1.0\linewidth}{!}{
\begin{tabular}{l|cc}
\toprule
Method (Regular Setting) & MAE $\downarrow$ & OBO $\uparrow$ \\ \hline
X3D~\cite{feichtenhofer2020x3d} & 0.9105 & 0.1059 \\
TANet~\cite{liu2021tam}   & 0.6624 & 0.0993 \\
Video SwinT~\cite{liu2021swin}   & 0.5756 & 0.1324 \\
Huang~\etal~\cite{huang2020improving}   & 0.5267 & 0.1589 \\
RepNet~\cite{dwibedi2020counting}   & 0.9950 & 0.0134 \\
Zhang~\etal~\cite{zhang2020context}   & 0.8786 & 0.1554 \\
TransRAC~\cite{hu2022transrac}   & 0.4431 & 0.2913 \\ 
MFL~\cite{li2024repetitive}$^\star$    & 0.3929 & 0.3444 \\ 
\rowcolor{gray!18}\textbf{Ours}   &\textbf{0.3543} & \textbf{0.3576} \\ \toprule
Method (Open-set Setting)  & MAE $\downarrow$ & OBO $\uparrow$ \\ \hline
Huang~\etal~\cite{huang2020improving}   & 1.0000  & 0.0000           \\
TransRAC~\cite{hu2022transrac}         & 0.6249            & 0.2040           \\ 
\rowcolor{gray!18}\textbf{Ours}  &  \textbf{0.5813} &    \textbf{0.3040} \\ 
\bottomrule
\end{tabular}
}
\label{tab:main}
\end{table}

\begin{table}[t!]
\centering
\tabcolsep 20pt
\caption{
Performance comparison on the UCFRep dataset. $^\dagger$: results obtained from MFL~\cite{li2024repetitive}.}
\resizebox{1.0\linewidth}{!}{
\begin{tabular}{l|cc}
\toprule
 Method & MAE $\downarrow$ & OBO $\uparrow$ \\ \hline
RepNet~\cite{dwibedi2020counting} & 0.9985 & 0.0090 \\
SightSound~\cite{zhang2021repetitive}$^\dagger$  & 0.4825 & 0.3125 \\ 
Zhang~\etal~\cite{zhang2020context}$^\dagger$ &  0.4689 & 0.4800 \\
TransRAC~\cite{hu2022transrac}$^\dagger$ & 0.4409 & 0.4300 \\ 
MFL~\cite{li2024repetitive}$^\dagger$ & 0.3879 & 0.5100 \\
\rowcolor{gray!18} \textbf{Ours} & \textbf{0.3483} & \textbf{0.5143} \\ \bottomrule  
\end{tabular}
}
\label{tab:ucfrep}
\end{table}

\subsection{Comparison with State-of-the-Arts}
\subsubsection{Intra-dataset Testing} 

\textbf{Results on RepCount-A.}
Following the protocol outlined in TransRAC~\cite{hu2022transrac}, we make comparisons with current state-of-the-art methods on the RepCount-A dataset under both \textbf{regular and open-set settings}. As shown in Table~\ref{tab:dataset}, we give the data statistics for the regular and open-set settings. 
In the regular setting, the video data is randomly divided, allowing all action categories to be present across the training, validation, and testing sets. Conversely, in the open-set setting, action categories are disjoint across the training, validation, and testing datasets, ensuring that the actions present in the test set have not been included in the training set. 
The open-set setting is particularly designed to assess the generalizability of the proposed model in handling action classes that were not encountered during training.

Compared with TransRAC~\cite{hu2022transrac}, our method not only generates diverse temporal-similarity matrices but also integrates local temporal contextual information for density map estimation. 
As shown in Table~\ref{tab:main}, in \textbf{the regular setting}, it is evident that our network outperforms TransRAC by a notable margin (\ie, MAE is reduced from 0.4431 to 0.3543, and OBO is improved from 0.2913 to 0.3576). In other words, the MAE drops down by 20.04\%, and the OBO is of 22.76\% improvement. 
Compared to the latest work MFL~\cite{li2024repetitive} that utilizes a reconstruction loss to strengthen the connection between two frames, our method also exhibits better performance, \ie, 0.3543 \vs 0.3929 in MAE, and 0.3576 \vs 0.3444 in OBO. Namely, MAE is improved by 9.82\%. 
As a repetitive action counting method, an important property is to count actions from unseen categories. 
In \textbf{the open-set setting}, the categories in train, validation, and test sets are completely different, which requires the model to have strong robustness. 
Note that category labels are not used in the model. 
As shown in Table~\ref{tab:main}, the proposed \M{} achieves the MAE and OBO of 0.5813 and 0.3040, respectively. 
Compared to TransRAC, our method improves the performance by 6.98\% and 49.02\% in terms of the MAE and OBO metrics, respectively.

\textbf{Results on UCFRep.} 
As shown in Table~\ref{tab:ucfrep}, we can see that our approach achieves state-of-the-art results in both MAE and OBO metrics, with an MAE of 0.3483 and an OBO of 0.5143. 
Compared to MFL~\cite{li2024repetitive}, which utilizes matrix reconstruction loss to enhance temporal correspondence between identical motion features, our method exhibits better performance by incorporating hybrid temporal relation modeling. Specifically, our method improves the MAE metric from 0.3879 to 0.3483, and increases the OBO metric from 0.5100 to 0.5143. These results further validate the effectiveness of our method in handling different repetitive action counting scenarios. 

\begin{table}[t]
\centering
\tabcolsep 5pt
\renewcommand\arraystretch{1.0}
\caption{Cross-dataset testing results of the UCFRep and QUVA Repetition datasets. The model is trained on the RepCount-A dataset. $^{\dag}$: re-implemented by our own using official code.} 
\resizebox{1.0\linewidth}{!}{
\begin{tabular}{l|cc|cc}
\toprule
\multirow{2}{*}{Method} & \multicolumn{2}{c|}{RepCount-A $\rightarrow$ UCFRep}  & \multicolumn{2}{c}{RepCount-A $\rightarrow$ QUVA} \\ \cline{2-5}
& MAE $\downarrow$   & OBO $\uparrow$ & MAE $\downarrow$   & OBO $\uparrow$ \\ 
\hline
RepNet~\cite{dwibedi2020counting}  & 0.9985 & 0.009 & 1.0000$^{\dag}$ & 0.000$^{\dag}$   \\
TransRAC~\cite{hu2022transrac} & 0.6401 & 0.324 & 0.5739$^{\dag}$& 0.110$^{\dag}$  \\ 
MFL~\cite{li2024repetitive} & 0.5227 & 0.350 & N/A & N/A \\ 
\rowcolor{gray!18} \textbf{Ours} & \textbf{0.5205} & \textbf{0.391}   & \textbf{0.5021} & \textbf{0.230} \\ 
\bottomrule
\end{tabular}}
\label{tab:transfer}
\end{table}

\subsubsection{Cross-dataset Testing} 
Here, we conduct cross-dataset experiments to evaluate the model's generalizability further. 
Specifically, we train the model on the RepCount-A dataset and evaluate the performance on the validation set of the UCFRep and QUVA Repetition datasets. 
\textbf{1) RepCount-A $\rightarrow$ UCFRep.} 
The UCFRep dataset has diverse time scales across various types of repetitions, accompanied by notable variations in the cycle lengths. 
As depicted in Table~\ref{tab:transfer}, we can observe that our method outperforms TransRAC with improvements of 18.68\% in MAE and 20.68\% in OBO. 
{}Compared to the la
These results further validate the superior generalizability of our model compared to TransRAC. 
\textbf{2) RepCount-A $\rightarrow$ QUVA Repetition.}
The QUVA Repetition dataset consists of 100 videos displaying a wide variety of repetitive video dynamics, such as swimming, stirring, and cutting. These categories are quite different from daily exercise activities in the RepCount-A dataset. 
From Table~\ref{tab:transfer}, we can see that our model achieves the best MAE of 0.5021 and OBO of 0.230. 
It is noticeable that the proposed method surpasses TransRAC by 12.51\% and 109.09\% in terms of MAE and OBO, respectively. 
In summary, these results further verify the generalization of \M{} in RAC among multiple datasets. 

\begin{table}[t]
\caption{Ablation study results of the main component on the RepCount-A dataset using RGB features.}
\renewcommand\arraystretch{1.0}
\resizebox{1.0\linewidth}{!}{
\begin{tabular}{l|cc}
\toprule
Method & MAE $\downarrow$  & OBO $\uparrow$ \\ \hline
 w/o local temporal context modeling & 0.4357 & 0.2715 \\ 
 w/o bi-modal temporal self-similarity modeling & 0.3856 & 0.3576  \\ 
 w/o random matrix dropping & 0.3723 & 0.3443 \\
 w/o multi-scale self-similarity matrix fusion & 0.4155 & 0.3377 \\ 
\rowcolor{gray!18} \textbf{\M{} (Ours)}& \textbf{0.3543} & \textbf{0.3576} \\
\bottomrule
\end{tabular}
}
\label{tab:main_comp}
\end{table}

\subsection{Ablation Studies}\label{sec:abl}
In this section, we perform in-depth ablation studies on the RepCount-A dataset to investigate and discuss the main components of the proposed method. 
\subsubsection{Main Components}
To investigate the effectiveness of the main components, we look into the impact of each component separately. 
The experimental results on the RepCount-A dataset are reported in Table \ref{tab:main_comp}. 
If \M{} without the local temporal context modeling, the performance drops by a large margin (\eg, MAE is dropped from 0.3543 to 0.4357, and OBO dropped from 0.3576 to 0.2715). 
This indicates that relying only on bi-modal temporal self-similarity matrix modeling to exploit the relationship between actions is not enough for RAC. 
In contrast, if only considering local temporal context (\M{} w/o bi-modal temporal self-similarity modeling) for RAC, the performance also decreases (\eg, MAE from 0.3543 to 0.3856). These results verify the necessity of temporal self-similarity matrix for RAC. 
As analyzed in Sec.~\ref{sec:embedding}, the random matrix dropping is proposed to force the model to learn diverse temporal correlations for action counting. 
We can find that \M{} w/o random matrix dropping leads to significant performance degradation (\eg, MAE from 0.3543 to 0.3723), which verifies the effectiveness of RMD. 
The proposed temporal similarity matrix fusion aims to learn temporal correlation in the matrices of three scales. 
The performance of \M{} w/o self-similarity matrix fusion degrades by a large margin, \ie, MAE and OBO degrade to 0.4155 and 0.3377, respectively. 
Therefore, we can conclude that the fusion of multi-scale matrix is important for feature aggregation in RAC. 

\begin{table}[t]
\centering
\footnotesize
\tabcolsep 4pt
\renewcommand\arraystretch{1.0}
\caption{Ablation study of our model with different temporal self-similarity matrix and drop strategies on the RepCount-A dataset.}
\resizebox{1.0\linewidth}{!}{
\begin{tabular}{cc|cc|cc}
\toprule
\multicolumn{2}{c|}{Self-similarity Strategy}  &  \multicolumn{2}{c|}{Drop Strategy}                 & \multicolumn{2}{c}{RepCount-A} \\ \hline
Self-attention & Dual-softmax & Dropout & {RMD} & MAE $\downarrow$ & OBO $\uparrow$    \\ \hline
\ding{51} &  -   & \ding{51} & - & {0.4006} & {0.3046}  \\ 
\ding{51} &  -  &  - & \ding{51} & 0.3905 & 0.3311   \\ \hline
- & \ding{51} & \ding{51} & - & {0.3917} & {0.3112} \\ 
- & \ding{51} & - & \ding{51} & 0.3858 &  0.3311   \\ \hline
 \ding{51} &\ding{51} & \ding{51} & - & {0.3831} & {0.3509} \\ 
\rowcolor{gray!18}\ding{51} &\ding{51} & - & \ding{51} & \textbf{0.3543}& \textbf{0.3576}   \\ 
\bottomrule
\end{tabular}
}
\label{tab:corr}
\end{table}

\subsubsection{Analysis on Temporal Self-Similarity Matrix Construction}
In Section~\ref{sec:embedding}, we argue our bi-modal temporal self-similarity matrix construction strategy is effective in capturing the temporal correlations between two frames. 
To prove this point, we compare these two self-similarity matrix generation methods separately, \ie, multi-head self-attention mechanism and multi-head dual-softmax operation. 
In addition, we also evaluate the RMD operation in our model with the Dropout~\cite{srivastava2014dropout}.  
As shown in Table~\ref{tab:corr}, the approach of combined multi-head self-attention mechanism and multi-head dual-softmax reaches the best performance (\ie, MAE and OBO of 0.3543 and 0.3576, respectively). 
It is evident that none of the individual methods for generating the similarity matrix can achieve the best result. 
The combination of multi-head self-attention and multi-head dual-softmax operation achieves optimal performance. 
For the drop strategy, we evaluate the model using the Dropout strategy with a consistent drop ratio $p$. 
These results provide evidence that the RMD strategy outperforms Dropout across different combinations of self-similarity matrix generation strategies. 
We attribute this superiority to the fact that the Dropout performs element-wise dropping, which is not suitable for multi-channel temporal self-similarity modeling.

\begin{figure}[t!]
\centering
\subfloat{\includegraphics[width=0.495\linewidth]{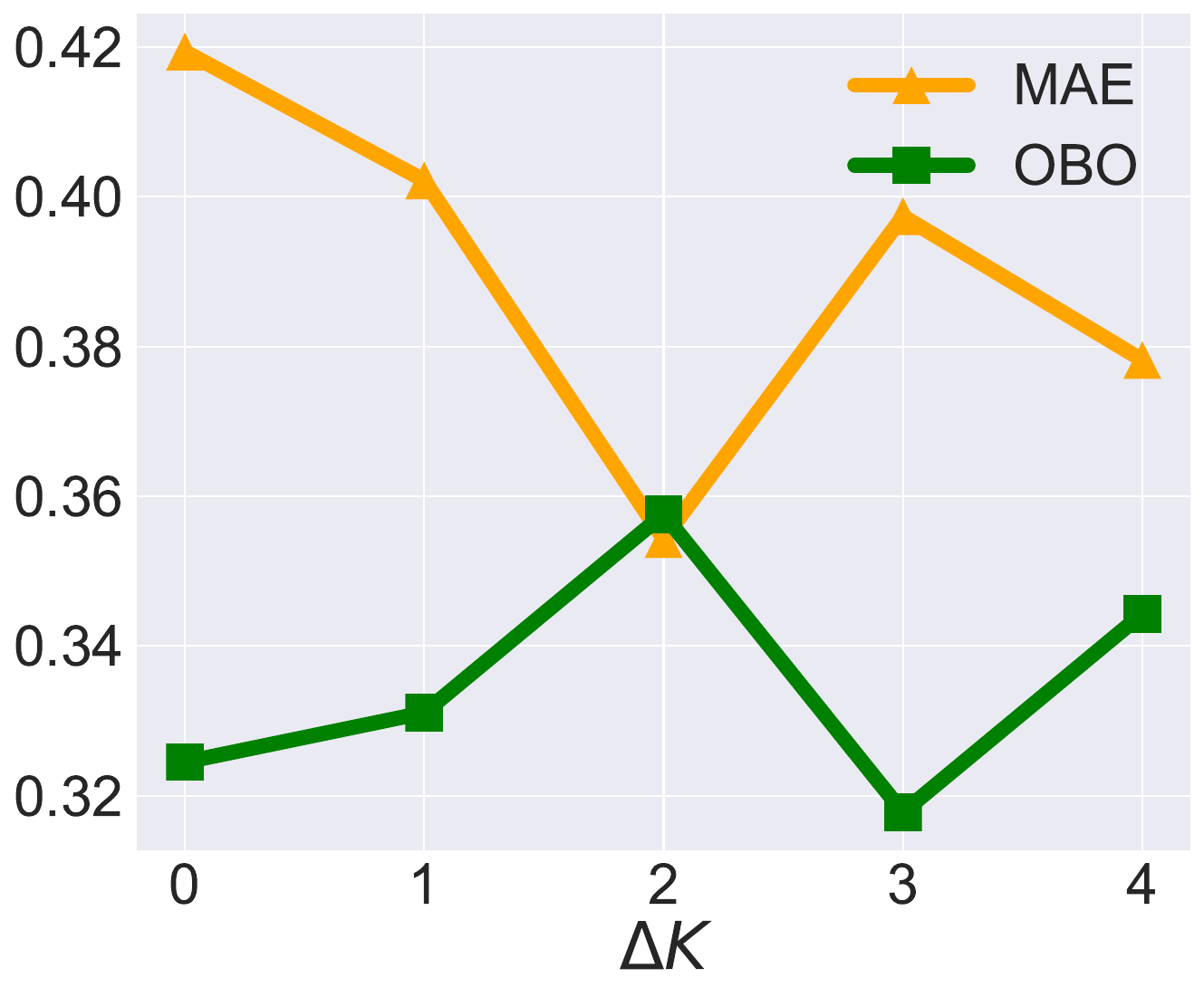}}
\subfloat{\includegraphics[width=0.495\linewidth]{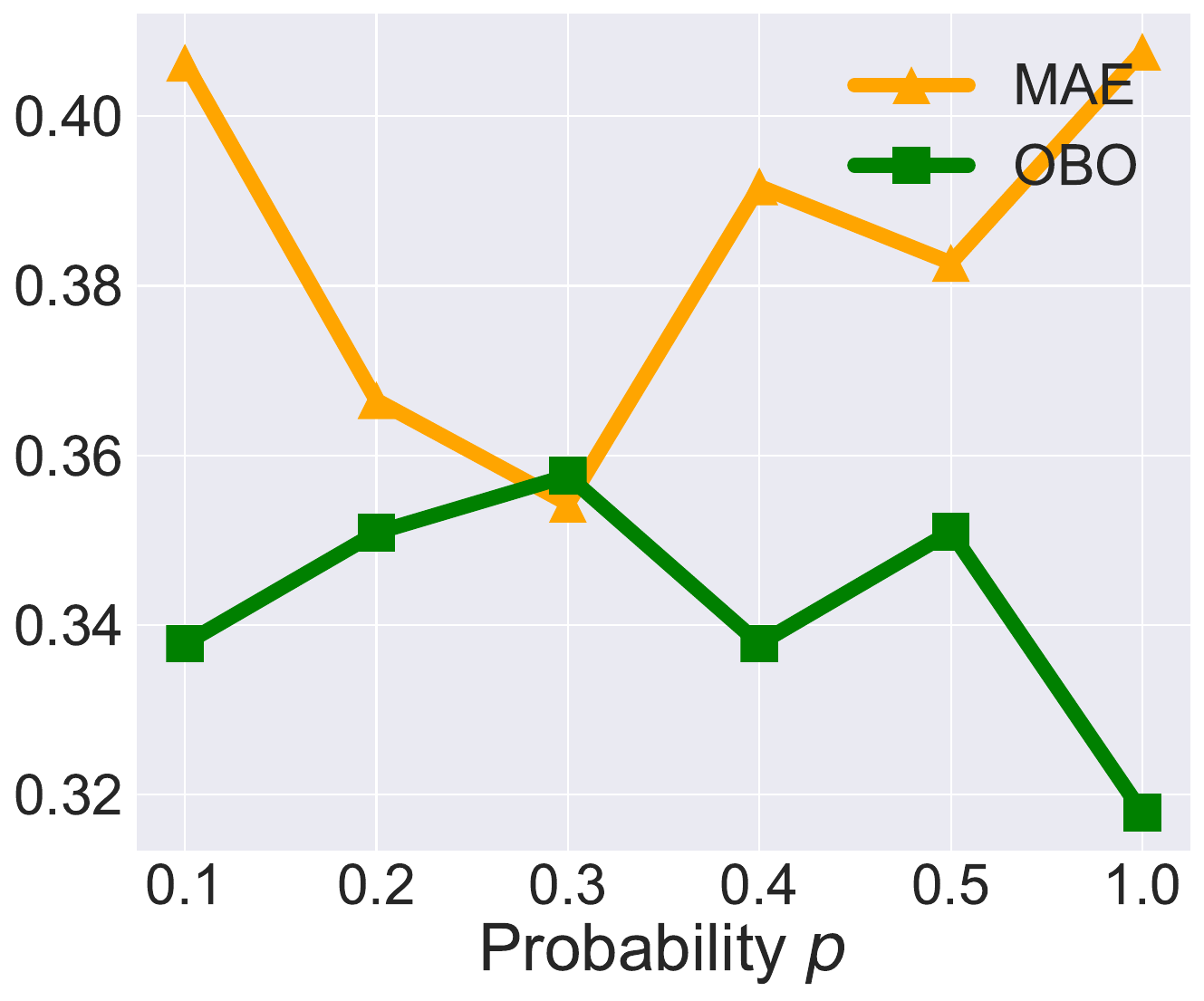}}
\caption{Ablation study of the drop ratio $p$ in the Random Matrix Dropping module and the parameter $\Delta K$ of the Local Temporal Context Modeling module on the RepCount-A dataset. $p$ changes with fixed $\Delta K$ = 2, and $\Delta K$ changes with fixed $p$ = 0.3.
}
\label{fig:pk}
\end{figure}

\begin{table}[t!]
\centering
\renewcommand\arraystretch{1.0}
\caption{Ablation results of different scale video features on RepCount-A. $\mathbf{V}_i, i\!\in\!\{1, 2, 3\}$ denotes different scale video feature.}
\tabcolsep 16pt
\resizebox{1.0\linewidth}{!}{
\begin{tabular}{c|c|cc}
\toprule
ID & Method  &  MAE $\downarrow$ & OBO $\uparrow$ \\ \hline
I & Ours-$\mathbf{V}_1$ & 0.5803   & 0.1987 \\
II & Ours-$\mathbf{V}_2$ & 0.4146   & 0.3113  \\
III & Ours-$\mathbf{V}_3$ & 0.4327   & 0.2914  \\ \hline
IV & Ours-$\mathbf{V}_{1,2}$ & 0.4037   & 0.3178 \\
V & Ours-$\mathbf{V}_{1,3}$ & 0.3914   & 0.3178  \\ \hline
\rowcolor{gray!18} VI &  \textbf{Ours} ($\mathbf{V}_{1,2,3}$) & \textbf{0.3543} & \textbf{0.3576} 
\\ \bottomrule
\end{tabular}}
\label{tab:ab_scale}
\end{table}

\subsubsection{Analysis on Random Matrix Dropping}
As shown in Figure~\ref{fig:pk}, we study the setting of drop ratio $p$ in RMD on the RepCount-A dataset. 
The larger the $p$, the higher the probability that the model will randomly drop the matrix during the training process. 
We can see that the model achieves the best performance when the value of $p$ is set to 0.3. 
Although some feature loss caused by random matrix dropping operation can be compensated by the local temporal context modeling, setting $p \textgreater 0.3$ leads to excessive information loss. On the contrary, setting $p\textless 0.3$ comes up with information redundancy. 
Therefore, we adopt $p$=0.3 as the optimal setting. 
\subsubsection{Analysis on Local Temporal Context Modeling}
To analyze the influence of local temporal context modeling, we evaluate the parameter $\Delta K$ of 1D separable convolution in Eq.~\ref{eq:conv1d} of the local temporal context modeling module to $\{0, 1, 2, 3, 4\}$. 
As shown in Figure~\ref{fig:pk}, we can see that the proposed \M{} achieves the best performance when $\Delta K=2$. 
With the increase of parameter $\Delta K$, the proposed \M{} will capture more local temporal contexts for density map estimation. 
However, too large receptive fields (\ie, $\Delta K \textgreater 2$) will result in capturing unnecessary and redundant contexts, which causes a decline in performance. 
Thus, we set $\Delta K = 2$ as the optimal setting. 

\subsubsection{Analysis on Multi-scale Video Features}
In this paper, we adopt the multi-scale video features for repetitive action counting.
This setting is a common practice in the RAC task~\cite{hu2022transrac,li2024repetitive}, targeting to accurately capture the dynamic patterns of high-frequency actions within non-uniform periods. 
Here, we conduct experiments to investigate different settings of multi-scale video features. 
The results are reported in Table~\ref{tab:ab_scale}. 
We can see that the single-scale features (models I, II, and III) perform the worst, while the two-scale features (models IV and V) exhibit slight improvement. 
In other words, the performance is improved by introducing multi-scale features. 
The best performance is achieved by model VI, which exploits the joint exploration of all the multi-scale features.

\begin{table}[t!]
\centering
\footnotesize
\tabcolsep 14pt
\renewcommand\arraystretch{1.0}
\caption{Ablation results of head number $H$ in temporal self-similarity generation on the RepCount-A dataset.}
\resizebox{1.0\linewidth}{!}{
\begin{tabular}{l|c|cc}
\toprule
Parameter & FLOPs $\downarrow$ & MAE $\downarrow$ & OBO $\uparrow$\\ \hline
$H$=2 & 146.54G & 0.3874 & 0.3112 \\ 
\textbf{$H$=4 (Ours)}& \textbf{146.55G} & \textbf{0.3543}  & \textbf{0.3576}  \\ 
$H$=8 &  146.57G & 0.3912  & 0.3178 \\ 
\bottomrule
\end{tabular}}
\label{tab:abl_head}
\end{table}

\begin{table}[t!]
\caption{Comparison of model complexity on the RepCount-A dataset under the regular setting. $\Delta$ denotes the relative improvements.}
\tabcolsep 3pt
\renewcommand\arraystretch{1.0}
\resizebox{1.0\linewidth}{!}{
\begin{tabular}{l|lll|ll}
\toprule
Method & \#params $\downarrow$ & FLOPs $\downarrow$ & Latency $\downarrow$& MAE $\downarrow$   & OBO $\uparrow$  \\ \hline
TransRAC~\cite{hu2022transrac} & 42.28M & 146.48G & 380.44ms & 0.4431 & 0.2913 \\ 
\textbf{Ours}     & 42.60M &   146.55G   & 382.70ms & \textbf{0.3543} & \textbf{0.3576} \\  
\rowcolor{gray!18} $\Delta$ & $+0.76\%$ & $+0.05\%$ & $+0.59\%$ &\textbf{+20.04\%} & \textbf{+22.76\%} \\ \bottomrule
\end{tabular}
}
\label{tab:params_cmp}
\end{table}

\begin{figure*}[t!]
\centering
\includegraphics[width=1.0\linewidth]{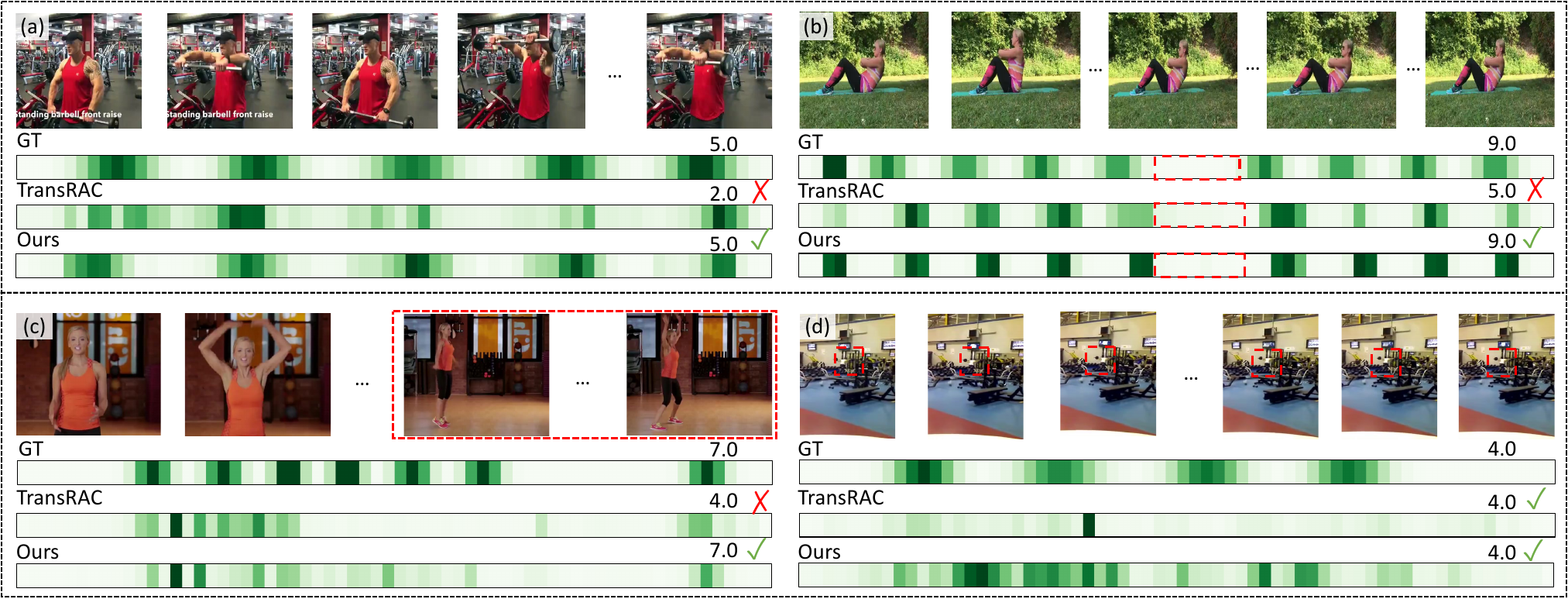}
\caption{Visualization of predicted density map on the RepCount-A dataset. (a), (b), and (c) are success cases, while (d) is a challenging case. 
In case (c), the latter frames of the video contain the action interruptions and viewpoint changes. 
``GT'' denotes the density map of ground truth. 
``TransRAC'' and ``Ours'' denote the prediction results of TransRAC~\cite{hu2022transrac} and our proposed method, respectively.} 
\label{fig:vis_dmap}
\end{figure*}

\subsubsection{Analysis on the Number of TSSM} 
In this paper, we utilize 4 heads (\ie, $H=4$) to generate TSSM in Eqs.~\ref{eq:m_sa} and~\ref{eq:m_ds}. 
The setting of 4 heads is a common practice and widely used in the task of repetitive action counting~\cite{hu2022transrac,li2024repetitive}. 
As depicted in Table~\ref{tab:abl_head}, we give the ablation results of $H\in\{2,4,8\}$ on the RepCount-A dataset. 
We can see that the model achieves the best performance when $H=4$. 
If the number of heads is increased, the performance will drop. In addition, more heads will lead to more assumptions of computational resources (\eg, FLOPs is increased from 146.55G to 146.57G when head number $H$ increases from 4 to 8). 
In order to balance the counting accuracy and inference efficiency, we use the setting of $H=4$ as the optimal choice.

\begin{figure}[t]
\centering
\includegraphics[width=1.0\linewidth]{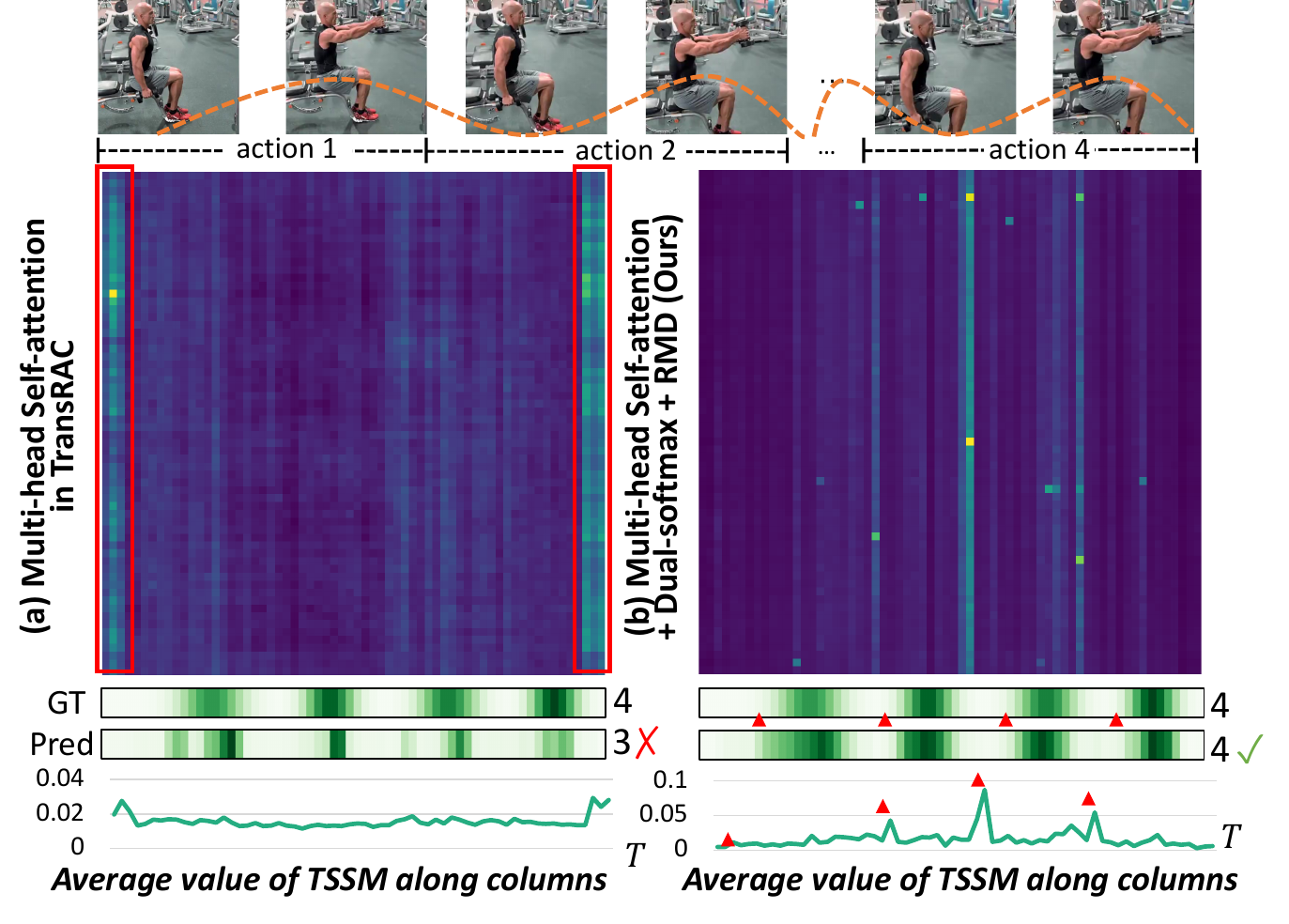}
\caption{
The illustration of the Temporal Self-Similarity Matrix (TSSM) in TransRAC (a) and our method (b) on the RepCount-A dataset. 
The TSSM in our method can capture the periodicity of repetitive actions. \textcolor{red}{$\blacktriangle$} denotes the starting of action.}
\label{fig:vis_ma}
\end{figure}

\subsubsection{Analysis on Model Complexity.}
To analyze the complexity of our model, we conduct experiments and report the size of parameters, FLOPs and inference latency in Table~\ref{tab:params_cmp}. 
Our method advances the Temporal Self-Similarity Matrix (TSSM) by incorporating hybrid temporal relation modeling, leading to a modest increase in parameters and computational load (\eg, the FLOPs (Floating Point Operations) are increased by 0.05\%). 
Considering that FLOPs serve as a metric for estimating the computational complexity of neural network architectures, they do not always accurately reflect execution speed on a GPU. 
While FLOPs offer an estimation of computational load, the real-time performance is influenced by multiple factors, including hardware optimization, network architecture, and memory access patterns. 
To this end, we give the average inference latency of our model and TransRAC, both evaluated on the same machine. 
We can see that our model only incurs a minimal increase in latency, amounting to 0.59\%  in milliseconds. 
Intuitively, our model is comparable to those of TransRAC~\cite{hu2022transrac}. 
Despite this operational similarity, our model demonstrates significant enhancements in performance, achieving improvements of 20.04\% and 22.76\% in MAE and OBO, respectively. These results underscore the efficiency and effectiveness of our approach.

\subsection{Qualitative Visualization and Analysis}
In this section, we visualize the learning temporal self-similarity matrix, learned features, and predicted density map on the RepCount-A dataset for qualitative analysis of the proposed \M{}. 
\subsubsection{Visualization of Temporal Self-Similarity Matrix} 
In this paper, we aim to build the diverse temporal self-similarity matrix to better capture the periodic information within the repetitive videos. 
As depicted in Figure~\ref{fig:vis_ma}, we illustrate the learned temporal self-similarity matrix, predicted density map, and the average value of TSSM along the column. 
The TSSM from TransRAC exhibits a clustered distribution in the left and right regions (marked in the red rectangle), which leads to inaccurate counts.  
In contrast, the TSSM generated by our method can capture the periodicity of actions, and there is a mapping from the curve peaks to the beginning of actions (marked in \textcolor{red}{$\blacktriangle$}). 
In summary, the self-similarity matrix constructed by our method is more informative than the self-similarity matrix generated by TransRAC.

\begin{figure}[t!]
\centering
\includegraphics[width=1.0\linewidth]{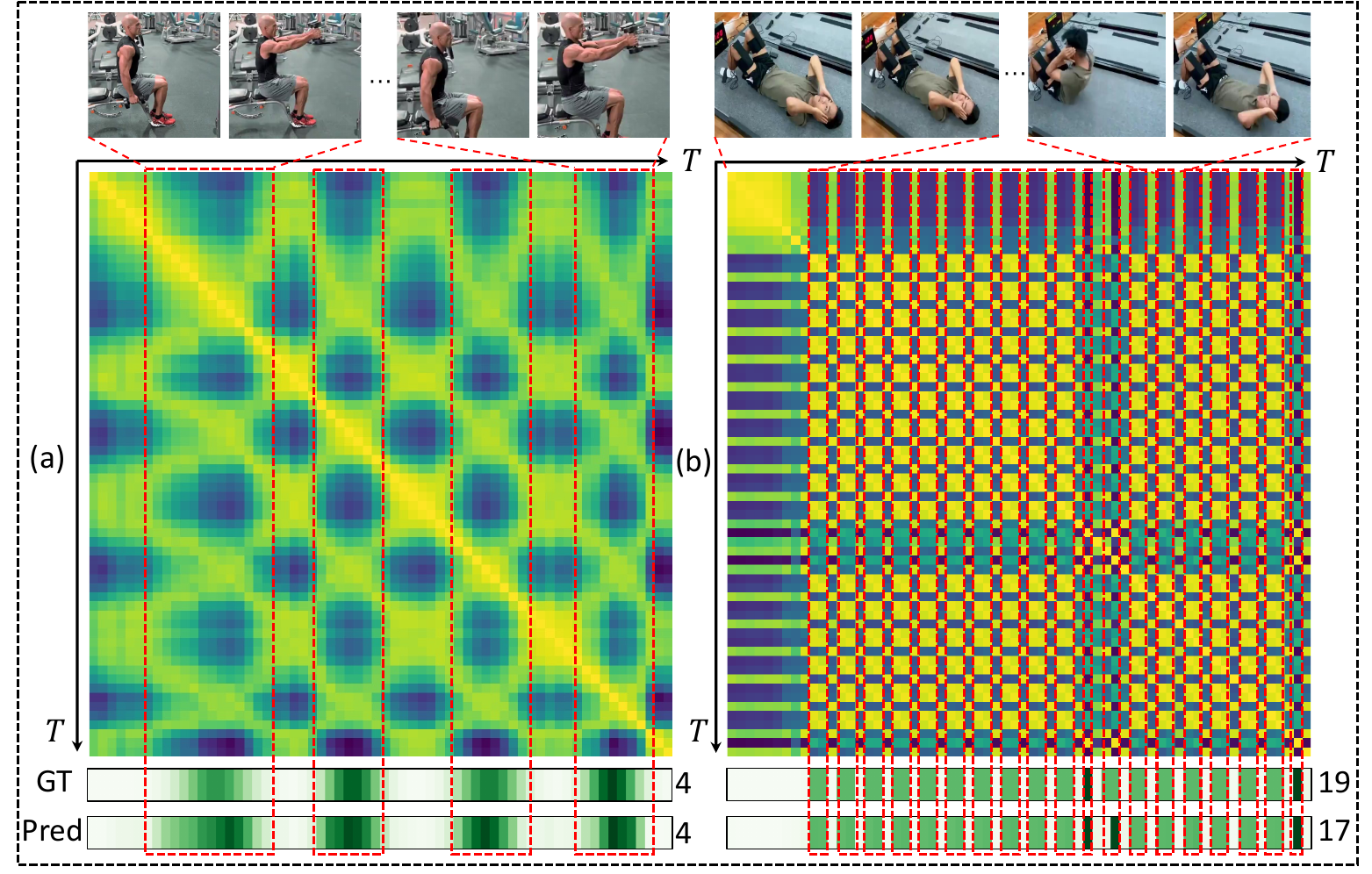}
\caption{Visualization of pairwise 
similarity score of the learned feature $\mathbf{O}$. The brighter color indicates a closer distance. 
Based on the hybrid temporal relation modeling, our method can extract the informative clues, \ie, the contents of actions (in red dashed line), which are helpful for the density map generation.
}
\label{fig:feature}
\end{figure}

\subsubsection{Visualization of Learned Features}
To verify the effectiveness of our method in capturing periodic information, we visually analyze the learned feature to verify the effectiveness of the proposed method. 
Specifically, we calculate the pairwise cosine similarity $\mathbf{M}_{cos} \in \mathbb{R}^{T\times T}$ of the learned feature $\mathbf{O} \in \mathbb{R}^{T\times d}$ in Eq.~\ref{eq:feature_O}, and then display it in Figure~\ref{fig:feature}. 
For instance, in example (a), the similarity matrix displays distinct blocks that align with the ground truth label, indicating that our method successfully extracts informative clues (\ie, the period patterns marked in red dashed rectangles) for density map estimation. 
Similarly, example (b) involves the repetitive actions that occur at high frequency. 
The visualizations from the temporal matrices and density map illustrate that our method adeptly captures informative clues and yields accurate counts. 
These findings demonstrate that our method can effectively aggregate informative clues from local temporal contexts and temporal self-similarity matrices to ensure accurate repetitive action counting.

\subsubsection{Visualization of Predicted Density Map} 
To demonstrate the effectiveness of the proposed \M{} for RAC, we visualize the predicted density map in Figure~\ref{fig:vis_dmap}. 
Concretely, case (a) illustrates the non-uniform periods, case (b) highlights the action interruptions in action sequences, and case (c) displays the repetitive actions with viewpoint changes. 
Our model demonstrates the robust capability to precisely predict the exact number of repetitions in these scenarios. 
Conversely, while TransRAC can predict the periods, it falls short in providing accurate densities of actions, resulting in imprecise counts. 
For instance, TransRAC predicts only 2 actions in case (a), and 5 actions in case (b). Our model not only successfully identifies the action periods in case (a) but also accurately detects the interruptions in case (b), yielding accurate and reliable counts. 
For case (c), we can observe that there are action interruptions and viewpoint changes (in red dashed box) in the last frames of the video. Our model can adapt the viewpoint changes and give truth counts, while TransRAC only predicts 4 actions in this case. 

Additionally, we also give a challenging sample in case (d), where the video contains a person performing squat exercises.
In this instance, the individual occupies a relatively small portion of the entire video frame. 
Interestingly, both our method and TransRAC successfully produce correct counts. However, TransRAC merely predicts one high density. 
In contrast, our method generates multiple high-density regions but falls short of precisely predicting the density map. 
In summary, this challenging case highlights that accurately counting repetitive actions of small visual objects remains a significant challenge in current research. 

\section{Conclusion}
In this paper, we presented a novel method named Hybrid Temporal Relation Modeling Network (\M{}) for repetitive action counting. 
We primarily focused on enhancing the representation of temporal self-similarity matrices through the exploitation of hybrid temporal relation modeling, including the bi-modal (\ie, multi-head self-attention and multi-head dual-softmax operations) temporal self-similarity modeling and local temporal contexts modeling. 
In addition, we also designed a random matrix dropping module that compels the model to learn rich information during the generation of temporal self-similarity matrix by randomly dropping the matrix. 
To capture the periodic information within each multi-scale matrix, we designed a multi-scale self-similarity fusion module to aggregate these matrices. 
Extensive experiments conducted on intra-dataset and cross-dataset demonstrate that our method achieves state-of-the-art results on all datasets. The qualitative visualization also shows the superiority of our method. 

{
\small
\bibliographystyle{IEEEtran}
\bibliography{IEEEtrans}

\begin{thebibliography}{10}
\providecommand{\url}[1]{#1}
\csname url@samestyle\endcsname
\providecommand{\newblock}{\relax}
\providecommand{\bibinfo}[2]{#2}
\providecommand{\BIBentrySTDinterwordspacing}{\spaceskip=0pt\relax}
\providecommand{\BIBentryALTinterwordstretchfactor}{4}
\providecommand{\BIBentryALTinterwordspacing}{\spaceskip=\fontdimen2\font plus
\BIBentryALTinterwordstretchfactor\fontdimen3\font minus
  \fontdimen4\font\relax}
\providecommand{\BIBforeignlanguage}[2]{{%
\expandafter\ifx\csname l@#1\endcsname\relax
\typeout{** WARNING: IEEEtran.bst: No hyphenation pattern has been}%
\typeout{** loaded for the language `#1'. Using the pattern for}%
\typeout{** the default language instead.}%
\else
\language=\csname l@#1\endcsname
\fi
#2}}
\providecommand{\BIBdecl}{\relax}
\BIBdecl

\bibitem{dwibedi2020counting}
D.~Dwibedi, Y.~Aytar, J.~Tompson, P.~Sermanet, and A.~Zisserman, ``Counting out
  time: Class agnostic video repetition counting in the wild,'' in
  \emph{Proceedings of the IEEE/CVF Conference on Computer Vision and Pattern
  Recognition}, 2020, pp. 10\,387--10\,396.

\bibitem{zhang2020context}
H.~Zhang, X.~Xu, G.~Han, and S.~He, ``Context-aware and scale-insensitive
  temporal repetition counting,'' in \emph{Proceedings of the IEEE/CVF
  Conference on Computer Vision and Pattern Recognition}, 2020, pp. 670--678.

\bibitem{hu2022transrac}
H.~Hu, S.~Dong, Y.~Zhao, D.~Lian, Z.~Li, and S.~Gao, ``Transrac: Encoding
  multi-scale temporal correlation with transformers for repetitive action
  counting,'' in \emph{Proceedings of the IEEE/CVF Conference on Computer
  Vision and Pattern Recognition}, 2022, pp. 19\,013--19\,022.

\bibitem{li2024repetitive}
X.~Li and H.~Xu, ``Repetitive action counting with motion feature learning,''
  in \emph{Proceedings of the IEEE/CVF Winter Conference on Applications of
  Computer Vision}, 2024, pp. 6499--6508.

\bibitem{xia2023exploring}
K.~Xia, L.~Wang, Y.~Shen, S.~Zhou, G.~Hua, and W.~Tang, ``Exploring action
  centers for temporal action localization,'' \emph{IEEE Transactions on
  Multimedia}, 2023.

\bibitem{wang2024low}
Y.~Wang, F.~Wang, K.~Li, X.~Feng, W.~Hou, L.~Liu, L.~Chen, Y.~He, and Y.~Wang,
  ``Low-light wheat image enhancement using an explicit inter-channel sparse
  transformer,'' \emph{Computers and Electronics in Agriculture}, vol. 224, p.
  109169, 2024.

\bibitem{tang2021graph}
S.~Tang, D.~Guo, R.~Hong, and M.~Wang, ``Graph-based multimodal sequential
  embedding for sign language translation,'' \emph{IEEE Transactions on
  Multimedia}, vol.~24, pp. 4433--4445, 2021.

\bibitem{li2023data}
K.~Li, D.~Guo, G.~Chen, F.~Liu, and M.~Wang, ``Data augmentation for human
  behavior analysis in multi-person conversations,'' in \emph{Proceedings of
  the 31st ACM International Conference on Multimedia}, 2023, pp. 9516--9520.

\bibitem{he2021dense}
T.~He, X.~Jin, X.~Shen, J.~Huang, Z.~Chen, and X.-S. Hua, ``Dense interaction
  learning for video-based person re-identification,'' in \emph{Proceedings of
  the IEEE/CVF International Conference on Computer Vision}, 2021, pp.
  1490--1501.

\bibitem{xu2024regennet}
L.~Xu, Y.~Zhou, Y.~Yan, X.~Jin, W.~Zhu, F.~Rao, X.~Yang, and W.~Zeng,
  ``Regennet: Towards human action-reaction synthesis,'' in \emph{Proceedings
  of the IEEE/CVF Conference on Computer Vision and Pattern Recognition}, 2024,
  pp. 1759--1769.

\bibitem{ran2007pedestrian}
Y.~Ran, I.~Weiss, Q.~Zheng, and L.~S. Davis, ``Pedestrian detection via
  periodic motion analysis,'' \emph{International Journal of Computer Vision},
  vol.~71, pp. 143--160, 2007.

\bibitem{guo2024benchmarking}
D.~Guo, K.~Li, B.~Hu, Y.~Zhang, and M.~Wang, ``Benchmarking micro-action
  recognition: Dataset, methods, and applications,'' \emph{IEEE Transactions on
  Circuits and Systems for Video Technology}, vol.~34, no.~7, pp. 6238--6252,
  2024.

\bibitem{liu2015novel}
H.~Liu, T.~Xu, X.~Wang, and Y.~Qian, ``A novel multi-feature descriptor for
  human detection using cascaded classifiers in static images,'' \emph{Journal
  of Signal Processing Systems}, vol.~81, pp. 377--388, 2015.

\bibitem{li2018structure}
X.~Li, H.~Li, H.~Joo, Y.~Liu, and Y.~Sheikh, ``Structure from recurrent motion:
  From rigidity to recurrency,'' in \emph{Proceedings of the IEEE Conference on
  Computer Vision and Pattern Recognition}, 2018, pp. 3032--3040.

\bibitem{qian2024dual}
W.~Qian, D.~Guo, K.~Li, X.~Zhang, X.~Tian, X.~Yang, and M.~Wang, ``Dual-path
  tokenlearner for remote photoplethysmography-based physiological measurement
  with facial videos,'' \emph{IEEE Transactions on Computational Social
  Systems}, vol.~11, pp. 4465--4477, 2024.

\bibitem{cutler2000robust}
R.~Cutler and L.~S. Davis, ``Robust real-time periodic motion detection,
  analysis, and applications,'' \emph{IEEE Transactions on Pattern Analysis and
  Machine Intelligence}, vol.~22, no.~8, pp. 781--796, 2000.

\bibitem{burghouts2006quasi}
G.~J. Burghouts and J.-M. Geusebroek, ``Quasi-periodic spatiotemporal
  filtering,'' \emph{IEEE Transactions on Image Processing}, vol.~15, no.~6,
  pp. 1572--1582, 2006.

\bibitem{briassouli2007extraction}
A.~Briassouli and N.~Ahuja, ``Extraction and analysis of multiple periodic
  motions in video sequences,'' \emph{IEEE Transactions on Pattern Analysis and
  Machine Intelligence}, vol.~29, no.~7, pp. 1244--1261, 2007.

\bibitem{pogalin2008visual}
E.~Pogalin, A.~W. Smeulders, and A.~H. Thean, ``Visual quasi-periodicity,'' in
  \emph{2008 IEEE Conference on Computer Vision and Pattern Recognition}, 2008,
  pp. 1--8.

\bibitem{levy2015live}
O.~Levy and L.~Wolf, ``Live repetition counting,'' in \emph{Proceedings of the
  IEEE International Conference on Computer Vision}, 2015, pp. 3020--3028.

\bibitem{zhang2021repetitive}
Y.~Zhang, L.~Shao, and C.~G. Snoek, ``Repetitive activity counting by sight and
  sound,'' in \emph{Proceedings of the IEEE/CVF Conference on Computer Vision
  and Pattern Recognition}, 2021, pp. 14\,070--14\,079.

\bibitem{li2018repetitive}
X.~Li, V.~Singh, Y.~Wu, K.~Kirchberg, J.~Duncan, and A.~Kapoor, ``Repetitive
  motion estimation network: Recover cardiac and respiratory signal from
  thoracic imaging,'' \emph{arXiv preprint arXiv:1811.03343}, 2018.

\bibitem{vaswani2017attention}
A.~Vaswani, N.~Shazeer, N.~Parmar, J.~Uszkoreit, L.~Jones, A.~N. Gomez,
  {\L}.~Kaiser, and I.~Polosukhin, ``Attention is all you need,''
  \emph{Advances in Neural Information Processing Systems}, vol.~30, 2017.

\bibitem{liu2022video}
Z.~Liu, J.~Ning, Y.~Cao, Y.~Wei, Z.~Zhang, S.~Lin, and H.~Hu, ``Video swin
  transformer,'' in \emph{Proceedings of the IEEE/CVF Conference on Computer
  Vision and Pattern Recognition}, 2022, pp. 3202--3211.

\bibitem{gong2024adaptive}
S.~Gong, J.~Yang, and S.~Zhang, ``Adaptive teaching for cross-domain crowd
  counting,'' \emph{IEEE Transactions on Multimedia}, vol.~26, pp. 2943--2952,
  2024.

\bibitem{zhu2024find}
H.~Zhu, J.~Yuan, X.~Zhong, L.~Liao, and Z.~Wang, ``Find gold in sand:
  Fine-grained similarity mining for domain-adaptive crowd counting,''
  \emph{IEEE Transactions on Multimedia}, vol.~26, pp. 3842--3855, 2024.

\bibitem{liang2023focal}
D.~Liang, W.~Xu, Y.~Zhu, and Y.~Zhou, ``Focal inverse distance transform maps
  for crowd localization,'' \emph{IEEE Transactions on Multimedia}, vol.~25,
  pp. 6040--6052, 2023.

\bibitem{guo2019dadnet}
D.~Guo, K.~Li, Z.-J. Zha, and M.~Wang, ``Dadnet: Dilated-attention-deformable
  convnet for crowd counting,'' in \emph{Proceedings of the 27th ACM
  international conference on multimedia}, 2019, pp. 1823--1832.

\bibitem{runia2018real}
T.~F. Runia, C.~G. Snoek, and A.~W. Smeulders, ``Real-world repetition
  estimation by div, grad and curl,'' in \emph{Proceedings of the IEEE
  Conference on Computer Vision and Pattern Recognition}, 2018, pp. 9009--9017.

\bibitem{chetverikov2006motion}
D.~Chetverikov and S.~Fazekas, ``On motion periodicity of dynamic textures,''
  in \emph{British Machine Vision Conference}, vol.~1, 2006, pp. 167--176.

\bibitem{belongie2006structure}
S.~Belongie and J.~Wills, ``Structure from periodic motion,'' in \emph{Spatial
  Coherence for Visual Motion Analysis}, 2006, pp. 16--24.

\bibitem{tsai1994cyclic}
P.-S. Tsai, M.~Shah, K.~Keiter, and T.~Kasparis, ``Cyclic motion detection for
  motion based recognition,'' \emph{Pattern Recognition}, vol.~27, no.~12, pp.
  1591--1603, 1994.

\bibitem{thangali2005periodic}
A.~Thangali and S.~Sclaroff, ``Periodic motion detection and estimation via
  space-time sampling,'' in \emph{2005 Seventh IEEE Workshops on Applications
  of Computer Vision (WACV/MOTION'05)-Volume 1}, vol.~2, 2005, pp. 176--182.

\bibitem{davis2000categorical}
J.~Davis, A.~Bobick, and W.~Richards, ``Categorical representation and
  recognition of oscillatory motion patterns,'' in \emph{Proceedings IEEE
  Conference on Computer Vision and Pattern Recognition}, 2000, pp. 628--635.

\bibitem{kay2017kinetics}
W.~Kay, J.~Carreira, K.~Simonyan, B.~Zhang, C.~Hillier, S.~Vijayanarasimhan,
  F.~Viola, T.~Green, T.~Back, P.~Natsev \emph{et~al.}, ``The kinetics human
  action video dataset,'' \emph{arXiv preprint arXiv:1705.06950}, 2017.

\bibitem{nam2021zero}
J.~Nam, D.~Ahn, D.~Kang, S.~J. Ha, and J.~Choi, ``Zero-shot natural language
  video localization,'' in \emph{Proceedings of the IEEE/CVF International
  Conference on Computer Vision}, 2021, pp. 1470--1479.

\bibitem{wang2024eulermormer}
F.~Wang, D.~Guo, K.~Li, and M.~Wang, ``Eulermormer: Robust eulerian motion
  magnification via dynamic filtering within transformer,'' in
  \emph{Proceedings of the AAAI Conference on Artificial Intelligence},
  vol.~38, no.~6, 2024, pp. 5345--5353.

\bibitem{qian2024cluster}
W.~Qian, K.~Li, D.~Guo, B.~Hu, and M.~Wang, ``Cluster-phys: Facial clues
  clustering towards efficient remote physiological measurement,'' in
  \emph{Proceedings of the 32nd ACM International Conference on Multimedia},
  2024, p. 330–339.

\bibitem{wang2024frequency}
F.~Wang, D.~Guo, K.~Li, Z.~Zhong, and M.~Wang, ``Frequency decoupling for
  motion magnification via multi-level isomorphic architecture,'' in
  \emph{Proceedings of the IEEE/CVF Conference on Computer Vision and Pattern
  Recognition}, 2024, pp. 18\,984--18\,994.

\bibitem{he2023transvcl}
S.~He, Y.~He, M.~Lu, C.~Jiang, X.~Yang, F.~Qian, X.~Zhang, L.~Yang, and
  J.~Zhang, ``Transvcl: attention-enhanced video copy localization network with
  flexible supervision,'' in \emph{Proceedings of the AAAI Conference on
  Artificial Intelligence}, vol.~37, no.~1, 2023, pp. 799--807.

\bibitem{liu2024conditional}
D.~Liu, J.~Zhu, X.~Fang, Z.~Xiong, H.~Wang, R.~Li, and P.~Zhou, ``Conditional
  video diffusion network for fine-grained temporal sentence grounding,''
  \emph{IEEE Transactions on Multimedia}, vol.~26, pp. 5461--5476, 2024.

\bibitem{li2021proposal}
K.~Li, D.~Guo, and M.~Wang, ``Proposal-free video grounding with contextual
  pyramid network,'' in \emph{Proceedings of the AAAI Conference on Artificial
  Intelligence}, vol.~35, no.~3, 2021, pp. 1902--1910.

\bibitem{kim2023language}
D.~Kim, J.~Park, J.~Lee, S.~Park, and K.~Sohn, ``Language-free training for
  zero-shot video grounding,'' in \emph{Proceedings of the IEEE/CVF Winter
  Conference on Applications of Computer Vision}, 2023, pp. 2539--2548.

\bibitem{du2022fast}
Z.~Du, X.~Wang, G.~Zhou, and Q.~Wang, ``Fast and unsupervised action boundary
  detection for action segmentation,'' in \emph{Proceedings of the IEEE/CVF
  Conference on Computer Vision and Pattern Recognition}, 2022, pp. 3323--3332.

\bibitem{junejo2010view}
I.~N. Junejo, E.~Dexter, I.~Laptev, and P.~Perez, ``View-independent action
  recognition from temporal self-similarities,'' \emph{IEEE Transactions on
  Pattern Analysis and Machine Intelligence}, vol.~33, no.~1, pp. 172--185,
  2010.

\bibitem{sun2015exploring}
C.~Sun, I.~N. Junejo, M.~Tappen, and H.~Foroosh, ``Exploring sparseness and
  self-similarity for action recognition,'' \emph{IEEE Transactions on Image
  Processing}, pp. 2488--2501, 2015.

\bibitem{kwon2021learning}
H.~Kwon, M.~Kim, S.~Kwak, and M.~Cho, ``Learning self-similarity in space and
  time as generalized motion for video action recognition,'' in
  \emph{Proceedings of the IEEE/CVF International Conference on Computer
  Vision}, 2021, pp. 13\,065--13\,075.

\bibitem{zhang2020counterfactual}
Z.~Zhang, Z.~Zhao, Z.~Lin, X.~He \emph{et~al.}, ``Counterfactual contrastive
  learning for weakly-supervised vision-language grounding,'' \emph{Advances in
  Neural Information Processing Systems}, vol.~33, pp. 18\,123--18\,134, 2020.

\bibitem{tan2021logan}
R.~Tan, H.~Xu, K.~Saenko, and B.~A. Plummer, ``Logan: Latent graph co-attention
  network for weakly-supervised video moment retrieval,'' in \emph{Proceedings
  of the IEEE/CVF Winter Conference on Applications of Computer Vision}, 2021,
  pp. 2083--2092.

\bibitem{ioffe2015batch}
S.~Ioffe and C.~Szegedy, ``Batch normalization: Accelerating deep network
  training by reducing internal covariate shift,'' in \emph{International
  Conference on Machine Learning}, 2015, pp. 448--456.

\bibitem{srivastava2014dropout}
N.~Srivastava, G.~Hinton, A.~Krizhevsky, I.~Sutskever, and R.~Salakhutdinov,
  ``Dropout: a simple way to prevent neural networks from overfitting,''
  \emph{The Journal of Machine Learning Research}, vol.~15, no.~1, pp.
  1929--1958, 2014.

\bibitem{chollet2017xception}
F.~Chollet, ``Xception: Deep learning with depthwise separable convolutions,''
  in \emph{Proceedings of the IEEE Conference on Computer Vision and Pattern
  Recognition}, 2017, pp. 1251--1258.

\bibitem{soomro2012ucf101}
K.~Soomro, A.~R. Zamir, and M.~Shah, ``Ucf101: A dataset of 101 human actions
  classes from videos in the wild,'' \emph{arXiv preprint arXiv:1212.0402},
  2012.

\bibitem{kingma2014adam}
D.~P. Kingma and J.~Ba, ``Adam: A method for stochastic optimization,''
  \emph{International Conference on Learning Representations}, 2014.

\bibitem{NEURIPS2019_9015}
A.~Paszke, S.~Gross, F.~Massa, A.~Lerer, J.~Bradbury, G.~Chanan, T.~Killeen,
  Z.~Lin, N.~Gimelshein, L.~Antiga, A.~Desmaison, A.~Kopf, E.~Yang, Z.~DeVito,
  M.~Raison, A.~Tejani, S.~Chilamkurthy, B.~Steiner, L.~Fang, J.~Bai, and
  S.~Chintala, ``Pytorch: An imperative style, high-performance deep learning
  library,'' in \emph{Advances in Neural Information Processing Systems 32},
  2019, pp. 8024--8035.

\bibitem{feichtenhofer2020x3d}
C.~Feichtenhofer, ``X3d: Expanding architectures for efficient video
  recognition,'' in \emph{Proceedings of the IEEE/CVF Conference on Computer
  Vision and Pattern Recognition}, 2020, pp. 203--213.

\bibitem{liu2021tam}
Z.~Liu, L.~Wang, W.~Wu, C.~Qian, and T.~Lu, ``Tam: Temporal adaptive module for
  video recognition,'' in \emph{Proceedings of the IEEE/CVF International
  Conference on Computer Vision}, 2021, pp. 13\,708--13\,718.

\bibitem{liu2021swin}
Z.~Liu, Y.~Lin, Y.~Cao, H.~Hu, Y.~Wei, Z.~Zhang, S.~Lin, and B.~Guo, ``Swin
  transformer: Hierarchical vision transformer using shifted windows,'' in
  \emph{Proceedings of the IEEE/CVF International Conference on Computer
  Vision}, 2021, pp. 10\,012--10\,022.

\bibitem{huang2020improving}
Y.~Huang, Y.~Sugano, and Y.~Sato, ``Improving action segmentation via
  graph-based temporal reasoning,'' in \emph{Proceedings of the IEEE/CVF
  Conference on Computer Vision and Pattern Recognition}, 2020, pp.
  14\,024--14\,034.

\end{thebibliography}
}

\end{document}